\documentclass[12pt,a4paper]{article}

\usepackage[T1]{fontenc}
\usepackage[utf8]{inputenc}
\usepackage{lmodern}

\usepackage[margin=2.5cm]{geometry}
\usepackage{setspace}
\usepackage{amsmath,amssymb,amsfonts}
\usepackage{mathtools}
\usepackage{bm}

\usepackage{graphicx}
\usepackage{booktabs}
\usepackage{multirow}
\usepackage{subcaption}
\usepackage{float}

\usepackage{framed}
\newcounter{algcounter}
\newenvironment{algorithm}[1][H]{%
  \refstepcounter{algcounter}%
  \begin{framed}\noindent\textbf{Algorithm \thealgcounter}%
}{%
  \end{framed}%
}
\newcommand{\REQUIRE}{\textbf{Require:} }
\newcommand{\FOR}[1]{\textbf{for} #1 \textbf{do}}
\newcommand{\ENDFOR}{\textbf{end for}}
\newcommand{\STATE}{\par\noindent\hspace{1em}}

\usepackage{xcolor}

\usepackage{hyperref}
\hypersetup{
    colorlinks=true,
    linkcolor=blue,
    citecolor=blue,
    urlcolor=blue
}

\usepackage[authoryear,round]{natbib}

\newcommand{\vect}[1]{\bm{#1}}

\begin{document}

% ---- Title ----
\title{\Large\bfseries Physics-Informed Neural Networks for Discovering Periodic Orbits
in the Gravitational Three-Body Problem}

\author{Nikolaos Kollias$^{*}$ \qquad Nikolaos Matzakos$^{\dagger}$}

\date{}
\maketitle

\renewcommand{\thefootnote}{\fnsymbol{footnote}}
\footnotetext[1]{School of Science and Technology,
Hellenic Open University, Athens, Greece.
E-mail: \texttt{n.kollias.v@gmail.com}.}
\footnotetext[2]{School of Science and Technology,
Hellenic Open University, Athens, Greece.
E-mail: \texttt{matzakos.nikolaos@ac.eap.gr}.
Also at School of Pedagogical and Technological Education (ASPETE),
Athens, Greece. E-mail: \texttt{nikmatz@aspete.gr}.}
\renewcommand{\thefootnote}{\arabic{footnote}}

% ---- Abstract ----
\begin{abstract}
Locating periodic solutions of chaotic dynamical systems normally
requires an initial guess close enough to the target orbit for numerical
continuation or gradient-based search to converge. We show that
Physics-Informed Neural Networks (PINNs) trained on sparse, noisy
observations \emph{without} initial conditions recover periodic orbits of
the gravitational three-body problem, including orbit families absent from
the training data. The method rests on a second-order ODE formulation,
fixed-frequency Fourier features, percentile-based adaptive refinement,
and a trainable scaling parameter, each validated on forward problems.
Across two 100-seed ensembles, $23$--$25\%$ of runs converge to families
not present in the training data.
We then ask what determines which family emerges. Two $\chi^2$ tests give a
consistent answer: changing the training data source significantly shifts
the distribution of recovered families ($p < 0.001$, Cram\'{e}r's
$V = 0.339$), whereas switching between the two initialization distributions
tested does not ($p = 0.620$, $V = 0.094$). The random seed selects which
family a given run recovers; the \emph{distribution} the weights are drawn
from does not shift the aggregate frequencies, but the training data does.
The evidence is empirical: we do not characterize the loss landscape
analytically, and PINNs remain slower than conventional integrators on
well-posed initial-value problems. What the experiments establish is that
the recovered orbits are verifiable rather than merely plausible: the
identified ones refine to genuine periodic solutions, a network trained on Lagrange data recovers the figure-eight
choreography (Li--Liao class I.A.1, matched to seven significant digits in
$T^*$), and one trained on figure-eight data recovers a
Broucke--Hadjidemetriou--H\'{e}non orbit closing to
$\delta_T < 10^{-9}$.
\end{abstract}


\bigskip

% ============================================================
\section{Introduction}
\label{sec:introduction}
% ============================================================

Finding periodic orbits of the gravitational three-body problem is a
search problem. Numerical continuation and gradient-based methods locate
them efficiently, but they require an initial guess close enough to the
target orbit for the iteration to converge---which is precisely what one
does not have when exploring an uncharted region of the solution space.
This paper asks whether a neural network can locate such orbits
\emph{without} that guess: from sparse, noisy observations alone, with no
initial conditions supplied.

Physics-Informed Neural Networks (PINNs) \citep{Raissi2019} make the
question well-posed. A PINN minimizes a composite loss combining a
data-fitting term with the residual of the governing equations; removing
the initial-condition constraint leaves an under-determined problem in
which the network must infer \emph{both} which orbit to follow and how to
follow it. We find that this problem is solvable, and that its solutions
are not unique: trained on the same noisy data with the same
hyperparameters, PINNs converge to physically distinct periodic orbit
families---Euler collinear, Lagrange equilateral, or
Broucke--Hadjidemetriou--H\'{e}non (BHH)---depending on the random seed,
and a substantial fraction of runs recover families \emph{absent from the
training data}.

This raises the natural follow-up question: what determines which
family a given run recovers? Two controlled $\chi^2$ tests separate the
seed from the sampling distribution and the training data
(Sections~\ref{subsec:statistical_robustness}
and~\ref{subsec:training_data_effect}).

\paragraph{The three-body problem as a testbed.}
The gravitational three-body problem is generically non-integrable
\citep{Poincare1892} and admits a rich landscape of periodic solutions:
from the classical Euler \citep{Euler1767} and Lagrange \citep{Lagrange1772}
families (often grouped as the Euler--Lagrange family in modern
classifications), to the figure-eight orbit---a choreography
\citep{Moore1993, Chenciner2000}---to the thousands of families now cataloged
\citep{Li2017catalog, Liao2022}.
This extraordinary multiplicity makes it an ideal testbed: there are
many orbits to be found, catalogs against which a claimed discovery can be
checked, and scale-invariant quantities ($T^*$, $L^*$) that constrain the
identification of a recovered orbit.

\paragraph{Related work.}
Early neural-network approaches for solving differential equations were
proposed by \citet{Lagaris1998}. The modern PINN framework, introduced by
\citet{Raissi2019}, faces well-documented
challenges including loss balancing \citep{Wang2021}, spectral bias
\citep{Rahaman2019, Wang2021FF}, and adaptive collocation
\citep{Wu2023, Karniadakis2021, Cuomo2022}.
Most relevant to our work, \citet{Zou2025} showed that PINNs can discover
multiple solutions to the same PDE through random initialization, though only
for reaction-diffusion and fluid flow with prescribed boundary conditions.
\citet{Matzakos2026} compare PINNs with Neural ODEs on the Morris--Lecar
neuron, finding PINNs more accurate in stiff regimes. Machine learning has
been applied to the $N$-body problem via surrogate models \citep{Breen2020}
and hybrid integrators \citep{SazUlibarrena2024}. \citet{Pereira2025}
recently applied a physics-regularized surrogate model to the three-body
problem as a forward solver, training on large trajectory datasets with known
initial conditions. Our work differs
fundamentally: we formulate the problem as an \emph{inverse} problem---training
on sparse, noisy observations \emph{without} initial conditions---and focus on
the multi-basin structure of the resulting loss landscape. To our
knowledge, no prior work has recovered distinct three-body periodic orbit
families with PINNs given no initial conditions, verified the recovered
orbits against published catalogs, or identified experimentally which
factor governs which family is recovered.

\paragraph{Contributions.}
Our central finding is that PINNs trained without initial conditions
recover periodic orbits of the three-body problem, including families
absent from the training data, and that the training data---not the weight
initialization---governs which families emerge. Specifically:

\begin{enumerate}
    \item \textbf{Orbit discovery without initial conditions (central
    result).} Given only sparse noisy observations and no initial
    conditions, PINNs recover periodic orbits across 200 seeds, with
    $23$--$25\%$ converging to families absent from the training data, and
    the inferred states refine to genuine periodic solutions---one matching
    the figure-eight choreography recovered from
    Lagrange data (Section~\ref{subsec:liliao_discovery}) and a BHH orbit
    from figure-eight data (Section~\ref{subsec:bhh_discovery}).

    \item \textbf{What controls basin selection.} Two controlled
    $\chi^2$ tests over three 100-seed ensembles show that the training data source
    significantly shifts the distribution of recovered families
    ($p < 0.001$, Cram\'{e}r's $V = 0.339$;
    Section~\ref{subsec:training_data_effect}), whereas the initialization
    distribution does not ($p = 0.620$, $V = 0.094$;
    Section~\ref{subsec:statistical_robustness}), identifying the data
    rather than the initialization scheme as the operative factor.

    \item \textbf{Methodological enablers (forward problem validation).}
    A second-order ODE formulation, fixed-frequency Fourier features,
    percentile-based RAR, and a trainable loss-balancing parameter
    enable accurate forward solutions that underpin the orbit discovery
    experiments (Table~\ref{tab:ablation_summary}).
\end{enumerate}

\paragraph{Scope.} This is a computational paper: the evidence is
empirical and statistical throughout. We do not prove that the loss
landscape contains multiple basins, nor characterize them analytically,
and the method is not a competitor to conventional integrators on
well-posed initial-value problems (Section~\ref{sec:discussion}). The claim
is narrower: PINNs can locate periodic orbits when no accurate initial
guess is available to seed a conventional search.

The paper is organized as follows. Section~\ref{sec:formulation} presents the
mathematical formulation. Section~\ref{sec:methodology} describes the PINN
methodology and technical innovations. Section~\ref{sec:experiments} summarizes
forward problem validation (detailed in Appendix~\ref{app:forward_details}).
Section~\ref{sec:orbit_discovery} presents the central results: orbit
discovery without initial conditions, the two $\chi^2$ tests isolating the
training data as the operative factor, and the cross-family discoveries
verified against published catalogs. Section~\ref{sec:discussion} places
these in the context of loss landscape theory, and
Section~\ref{sec:conclusions} concludes.

% ============================================================
\section{Mathematical formulation}
\label{sec:formulation}
% ============================================================

\subsection{Equations of motion}
\label{subsec:equations}

Consider three point masses $m_i$, $i = 1, 2, 3$, interacting solely through
Newtonian gravity. Let $\vect{r}_i = (x_i, y_i)$ denote the position of the
$i$-th body in the plane. The equations of motion are given by Newton's second law:
\begin{equation}
    \ddot{\vect{r}}_i = -G \sum_{\substack{j=1 \\ j \neq i}}^{3} m_j
    \frac{\vect{r}_i - \vect{r}_j}{|\vect{r}_i - \vect{r}_j|^3},
    \quad i = 1, 2, 3,
    \label{eq:newton}
\end{equation}
where $G$ is the gravitational constant. The restriction to planar motion is
justified when the initial positions and velocities are coplanar: the
gravitational forces then lie in the span of the position differences, so
the plane is invariant under the dynamics.

In component form, we obtain six second-order ODEs:
\begin{equation}
    \ddot{x}_i = -G \sum_{\substack{j=1 \\ j \neq i}}^{3} m_j
    \frac{x_i - x_j}{r_{ij}^3}, \quad
    \ddot{y}_i = -G \sum_{\substack{j=1 \\ j \neq i}}^{3} m_j
    \frac{y_i - y_j}{r_{ij}^3},
    \label{eq:components_2nd}
\end{equation}
where $r_{ij} = \sqrt{(x_i - x_j)^2 + (y_i - y_j)^2}$.

\subsubsection{Numerical regularization of close encounters}
\label{subsec:regularization}
During PINN training, the network outputs $\vect{r}_i(t)$ are not constrained to
remain well-separated, and transient configurations with $r_{ij} \to 0$ can arise
before convergence. To prevent numerical overflow in the gravitational force
evaluation, we regularize the pairwise distances by replacing $r_{ij}$ with
$\tilde{r}_{ij} = \sqrt{r_{ij}^2 + \varepsilon}$, where
$\varepsilon$ is a small softening parameter (effective softening
length $\sqrt{\varepsilon}$). We set $\varepsilon = 10^{-9}$
($\sqrt{\varepsilon} \approx 3.2 \times 10^{-5}$) throughout all
experiments. This ensures
that the loss function and its gradients remain finite throughout training without
affecting the converged solution, since at convergence $r_{ij} \gg \sqrt{\varepsilon}$
for all time steps.

\subsubsection{First-order formulation}
The standard numerical approach introduces velocity variables $v_{xi} = \dot{x}_i$,
$v_{yi} = \dot{y}_i$, yielding 12 first-order ODEs:
\begin{equation}
    \dot{x}_i = v_{xi}, \quad \dot{v}_{xi} = -G \sum_{\substack{j=1 \\ j \neq i}}^{3}
    m_j \frac{x_i - x_j}{r_{ij}^3}, \quad \text{(and similarly for } y_i\text{)}.
    \label{eq:first_order}
\end{equation}

\subsubsection{Non-dimensionalization}
Setting $G = 1$ and choosing the largest mass $M$ as the mass scale and an appropriate
length scale $L$ (e.g., the largest initial separation), we obtain a dimensionless
system with time scale $T_\text{char} = \sqrt{L^3/(GM)}$. For equal-mass problems
($m_1 = m_2 = m_3 = 1$), we additionally rescale time so that the orbital period
maps to a value within the effective range $[-2, 2]$ of the $\tanh$ activation
function (the importance of this step is demonstrated in
Appendix~\ref{app:lagrange_orbits}). The Newtonian three-body equations are
invariant under the scaling transformation
\begin{equation}
    \vect{r} \to \alpha\,\vect{r}, \quad
    t \to \alpha^{3/2}\,t, \quad
    \vect{v} \to \alpha^{-1/2}\,\vect{v},
    \label{eq:scaling_law}
\end{equation}
where $\alpha > 0$ is an arbitrary scaling factor
\citep{Li2017catalog}. Only the numerical values of the initial
conditions and period change; the equations of motion retain their form.

\subsection{Physics-Informed Neural Networks}
\label{subsec:pinns}

A PINN approximates the solution $\vect{u}(t)$ of a differential equation by a
neural network $\mathcal{N}(t; \vect{\theta})$ with parameters $\vect{\theta}$.
Training minimizes a composite loss function:
\begin{equation}
    \mathcal{L}(\vect{\theta}) = w_r \mathcal{L}_r(\vect{\theta})
    + w_{\text{ic}} \mathcal{L}_{\text{ic}}(\vect{\theta})
    + w_d \mathcal{L}_d(\vect{\theta}),
    \label{eq:loss_total}
\end{equation}
where $\mathcal{L}_r$ is the PDE/ODE residual loss evaluated at collocation points,
$\mathcal{L}_{\text{ic}}$ enforces initial conditions, $\mathcal{L}_d$ is the
data-fitting loss (when observational data are available), and $w_r, w_{\text{ic}}, w_d$
are weighting coefficients.

For the second-order system \eqref{eq:components_2nd}, the residual loss takes the form:
\begin{equation}
    \mathcal{L}_r = \frac{1}{N_r} \sum_{k=1}^{N_r} \sum_{i=1}^{3}
    \left[ \left( \ddot{x}_i^{\mathcal{N}}(t_k) - f_{xi}(t_k) \right)^2
    + \left( \ddot{y}_i^{\mathcal{N}}(t_k) - f_{yi}(t_k) \right)^2 \right],
    \label{eq:residual_loss}
\end{equation}
where $f_{xi}, f_{yi}$ denote the gravitational acceleration components computed
from the network outputs, and the second derivatives $\ddot{x}_i^{\mathcal{N}},
\ddot{y}_i^{\mathcal{N}}$ are obtained via automatic differentiation.

\subsection{Hard enforcement of initial conditions}
\label{subsec:hard_constraints}

Soft enforcement of initial conditions via penalty terms in the loss function often
fails for gravitational problems, where the $1/r^2$ force law produces large
accelerations when bodies approach each other. We instead enforce initial conditions
exactly through an output transformation:
\begin{equation}
    \vect{y}_\phi(t; \vect{\theta}) = \vect{y}_0 + t\,\vect{u}_0
    + t^2\,\mathcal{N}(t; \vect{\theta}),
    \label{eq:hard_constraint}
\end{equation}
where $\vect{y}_0$ and $\vect{u}_0$ are the initial positions and velocities,
respectively. This guarantees $\vect{y}_\phi(0) = \vect{y}_0$ and
$\dot{\vect{y}}_\phi(0) = \vect{u}_0$ identically, eliminating the need for
initial condition loss terms.

% ============================================================
\section{Methodology}
\label{sec:methodology}
% ============================================================

\subsection{Network architecture and training}
\label{subsec:architecture}

All experiments use fully connected feedforward networks with $\sin$
activation for data-driven orbit discovery experiments
(Section~\ref{sec:orbit_discovery}) and $\tanh$ for forward problems
(Section~\ref{sec:experiments}), implemented in the DeepXDE library
\citep{Lu2021} with TensorFlow backend. Architectures range from
$3 \times 64$ to $5 \times 64$ (hidden layers $\times$ neurons per layer),
with $3 \times 128$ for the Euler forward experiments.
Forward problems use a two-phase training strategy: Adam
\citep{Kingma2015} ($\eta = 10^{-4}$, $2$--$3 \times 10^5$ epochs) followed
by L-BFGS ($1.5 \times 10^4$ epochs). Data-driven experiments use AdamW
\citep{Loshchilov2019} with problem-specific settings.
Reference solutions are computed with \texttt{scipy.integrate.solve\_ivp}
($\texttt{rtol} = 10^{-10}$, $\texttt{atol} = 10^{-12}$).
Table~\ref{tab:hyperparameters} summarizes all hyperparameters.

\begin{table}[H]
\centering
\caption{Common hyperparameters across all experiments. Experiment-specific
deviations are noted in the relevant subsections.}
\label{tab:hyperparameters}
\begin{tabular}{lll}
\toprule
\textbf{Hyperparameter} & \textbf{Value} & \textbf{Notes} \\
\midrule
Activation function & $\tanh$ / $\sin$ & $\tanh$ forward; $\sin$ orbit discovery \\
Hidden layers & $3$--$5$ & Problem-dependent \\
Neurons per layer & $64$--$128$ & Problem-dependent \\
Adam learning rate & $10^{-4}$ & Forward problems; see Sec.~\ref{sec:orbit_discovery} for data-driven \\
Adam epochs & $2 \times 10^5$--$3 \times 10^5$ & \\
L-BFGS epochs & $1.5 \times 10^4$ & \\
Precision & 64-bit (float64) & \\
Collocation points & $64$ & Before RAR additions \\
Softening $\varepsilon$ & $10^{-9}$ & All experiments \\
Initialization & Glorot Uniform & Unless stated otherwise \\
\bottomrule
\end{tabular}
\end{table}

\subsection{First-order vs.\ second-order ODE formulations}
\label{subsec:order_comparison}

A key methodological question for PINNs applied to second-order systems is whether
to reduce to first order. The first-order system \eqref{eq:first_order} for three
bodies produces 12 residual loss terms (6 acceleration equations + 6 velocity
definitions), while the second-order system \eqref{eq:components_2nd} produces only
6 terms. Moreover, the velocity-definition residuals ($\dot{x}_i - v_{xi} = 0$)
and acceleration residuals typically have very different magnitudes, creating an
imbalanced multi-objective optimization problem. In contrast, the 6 second-order
residuals are all acceleration terms of comparable magnitude.

This motivates our preference for the second-order formulation, validated
experimentally in Appendix~\ref{app:euler_orbits} with all other
hyperparameters held identical.

\subsection{Fourier feature input transformation}
\label{subsec:fourier}

For the figure-eight orbit \citep{Chenciner2000}, where three equal-mass bodies
trace the same lemniscate curve with a temporal phase offset of $T/3$
($2\pi/3$ for the fundamental, $4\pi/3$ for the second harmonic),
standard PINNs struggle to resolve
the symmetric structure. We introduce a Fourier feature transformation of the
input:
\begin{equation}
\begin{split}
    \Phi(t) = \Big[\, t,\;
    &\sin(\omega t),\; \cos(\omega t),\;
    \sin\!\left(\omega t + \tfrac{2\pi}{3}\right),\;
    \cos\!\left(\omega t + \tfrac{2\pi}{3}\right), \\
    &\sin\!\left(2\omega t + \tfrac{2\pi}{3}\right),\;
    \cos\!\left(2\omega t + \tfrac{2\pi}{3}\right) \Big],
\end{split}
    \label{eq:fourier_features}
\end{equation}
where $\omega = 2\pi/T$ is the fundamental angular frequency. The network then
operates as $\mathcal{N}(\Phi(t); \vect{\theta})$, receiving 7 features instead
of 1.

Two points about this map deserve statement, since the motivation is
easily overstated. First, the phase offsets are not independent
information: $\sin(\omega t + 2\pi/3)$ and $\cos(\omega t + 2\pi/3)$ are
linear combinations of $\sin \omega t$ and $\cos \omega t$, so under an
affine first layer the shifted pair spans the same subspace as the
unshifted one. The map~\eqref{eq:fourier_features} has rank~5, not~7: it
supplies $t$ together with the fundamental and the second harmonic.
Second, it is therefore the presence of the harmonics---not the encoding
of the $T/3$ phase symmetry---that accounts for the improvement over a
plain time input. This is consistent with the measured results, where
adding the phase shifts to a plain Fourier encoding changes RMSE by a
factor of $2.7$ (Table~\ref{tab:figure_eight_comparison}) while removing
the Fourier features altogether costs three orders of magnitude.

\subsection{Modified Residual-based Adaptive Refinement (RAR)}
\label{subsec:rar}

Standard RAR \citep{Lu2021} iteratively adds the collocation points with the largest
residuals to the training set. We observe that this greedy strategy can lead to
overfitting at extreme outlier points without improving global accuracy. We propose
a modified RAR that at each iteration:
\begin{enumerate}
    \item Evaluates residuals on a large random candidate set ($N = 10{,}000$ points).
    \item Selects the subset with residuals between the 60th and 95th percentiles.
    \item Randomly samples $n_\text{add} = 64$ points from this subset.
    \item Adds the selected points to the training set and continues training.
\end{enumerate}

The 60th percentile lower bound excludes regions where the residual
is already well-resolved; the 95th percentile upper bound excludes extreme
outliers near gravitational singularities, which standard RAR
over-represents. This promotes better generalization than greedy
top-$k$ selection (see Algorithm~\ref{alg:modified_rar} and
Table~\ref{tab:lagrange_rar_comparison}).
The candidate set is drawn once and reused across all RAR iterations,
which is what the released code implements; resampling it at every
iteration produced no appreciable difference in our tests.

\begin{algorithm}[H]
\textbf{: Modified RAR for PINNs}\label{alg:modified_rar}

\smallskip
\REQUIRE Trained model $\mathcal{N}$, domain $\Omega$, candidate size $N$,
points to add $n_\text{add}$, percentile range $[p_\text{low}, p_\text{high}]$,
RAR iterations $K$, training epochs per iteration $E$
\STATE Sample a fixed candidate set $\{t_j\}_{j=1}^N$ uniformly from $\Omega$
\STATE \FOR{$k = 1$ to $K$}
\STATE \hspace{1em}1. Compute mean absolute residual $\bar{R}(t_j)$ for each $t_j$
\STATE \hspace{1em}2. $\tau_\text{low} \leftarrow \text{Percentile}(\{\bar{R}\}, p_\text{low})$
\STATE \hspace{1em}3. $\tau_\text{high} \leftarrow \text{Percentile}(\{\bar{R}\}, p_\text{high})$
\STATE \hspace{1em}4. $S \leftarrow \{t_j : \tau_\text{low} \leq \bar{R}(t_j) \leq \tau_\text{high}\}$
\STATE \hspace{1em}5. Randomly select $n_\text{add}$ points from $S$; add to training set
\STATE \hspace{1em}6. Train $\mathcal{N}$ for $E$ additional epochs
\STATE \ENDFOR
\end{algorithm}

\subsection{Trainable residual scaling parameter}
\label{subsec:trainable_C}

When training PINNs with observational data and without initial conditions, the PDE
residual gradients typically dominate the data-fitting gradients, especially early in
training when network outputs are far from the true solution. To address this imbalance,
we introduce a trainable parameter $C$ that scales each PDE residual by $1/C^2$
before squaring. Since the loss is a sum of squared residuals, this yields:
\begin{equation}
    \tilde{\mathcal{L}}_r = \frac{1}{C^4} \mathcal{L}_r,
    \label{eq:scaled_residual}
\end{equation}
where $C$ is initialized to $C_0 = 3$ and optimized
jointly with the network parameters.
This sits between two existing lines of work. Adaptive loss-weighting
schemes such as \citet{Wang2021} adjust the weights externally during
training, while \citet{Theodosiou2026} avoid loss balancing altogether by
scaling each term of the governing equations before the loss is assembled.
Our scaling is likewise applied to the residual rather than to the
assembled loss, but the factor is not fixed in advance: $C$ is an ordinary
trainable parameter optimized jointly with the network, so the schedule
emerges from training. The practical consequence is that no modification
to the training loop is required, and the method works within standard
PINN frameworks that expose only static loss weights. Since $\partial \tilde{\mathcal{L}}_r /
\partial C = -4 C^{-5} \mathcal{L}_r < 0$, the parameter $C$ increases monotonically
during training, progressively reducing the weight of the PDE residuals and allowing
the data-fitting loss to exert more influence as training progresses.

% ============================================================
\section{Forward problem experiments}
\label{sec:experiments}
% ============================================================

We validate our PINN methodology on forward problems---using PINNs
as solvers without external data, with hard-constrained initial
conditions---before turning to orbit discovery
(Section~\ref{sec:orbit_discovery}). Full experimental details, tables, and
figures are provided in Appendix~\ref{app:forward_details}; here we summarize
the key findings.

\paragraph{Two-body validation.}
\label{subsec:two_body}
A $3 \times 64$ PINN achieves position RMSE of $2.12 \times 10^{-7}$
on the Kepler two-body problem, confirming the accuracy of our framework.

\paragraph{Euler orbits: second-order formulation.}
\label{subsec:euler_orbits}
For Euler collinear orbits ($m_1 = m_2 = m_3 = 1$), the second-order ODE
formulation (6~equations) reduces the maximum position error by approximately
two orders of magnitude compared to the first-order system (12~equations),
from $1.28 \times 10^{-1}$ to $7.99 \times 10^{-4}$
(Table~\ref{tab:euler_comparison}). The imbalanced residual magnitudes in
the first-order system cause the ``stationary'' body~3 to drift from
its equilibrium, while the second-order system maintains stability.

\paragraph{Lagrange orbits: time rescaling and modified RAR.}
\label{subsec:lagrange_orbits}
Time rescaling to match the $\tanh$ activation's effective range improves RMSE
by a factor of ${\sim}\,45$. Fixed collocation with L-BFGS drives the training loss to
$4.33 \times 10^{-4}$ while the test loss reaches $1.31$---a gap of three
to four orders of magnitude. Our modified RAR strategy (60th--95th
percentile sampling) closes it, converging with train and test losses
within $10\%$ of one another
(Table~\ref{tab:lagrange_rar_comparison}).

\paragraph{Figure-eight orbits: Fourier features.}
\label{subsec:figure_eight}
Standard PINNs fail on the figure-eight orbit: test loss diverges despite
decreasing training loss. The Fourier feature
transformation~\eqref{eq:fourier_features}, which supplies the fundamental
and second harmonic of the known orbital period, achieves RMSE of $1.53 \times 10^{-4}$
(Table~\ref{tab:figure_eight_comparison}).

\paragraph{Conservation law diagnostics.}
\label{subsec:conservation}
The PINN solutions conserve energy to within $\sim\!0.02$--$3\%$ over one
orbital period (Table~\ref{tab:conservation}), which provides an error measure
independent of the training loss, since conservation is never enforced
explicitly.

\paragraph{Ablation summary.}
Table~\ref{tab:ablation_summary} summarizes the contribution of each
methodological component. Each row isolates one component against the
comparable configuration without it, on the orbit where its effect is most
pronounced.
\begin{table}[H]
\centering
\caption{Ablation summary: contribution of each methodological
component. Metric is position RMSE unless noted. Detailed results in the
referenced tables.}
\label{tab:ablation_summary}
\small
\begin{tabular}{llcccl}
\toprule
\textbf{Component} & \textbf{Orbit} & \textbf{Without} & \textbf{With} &
\textbf{Gain} & \textbf{Ref.} \\
\midrule
2nd-order ODE$^{\dagger}$ & Euler & $1.28 \times 10^{-1}$ & $7.99 \times 10^{-4}$ &
$\sim\!160\times$ & Tab.~\ref{tab:euler_comparison} \\
Fourier features & Fig.-eight & $2.50 \times 10^{-1}$ & $1.53 \times 10^{-4}$ &
$\sim\!1600\times$ & Tab.~\ref{tab:figure_eight_comparison} \\
Percentile RAR & Lagrange & $3.26 \times 10^{-2}$ & $1.05 \times 10^{-2}$ &
$\sim\!3\times$ & Tab.~\ref{tab:lagrange_rar_comparison} \\
Trainable $C$ & Lagrange$^*$ & $1.85 \times 10^{-1}$ & $7.98 \times 10^{-3}$ &
$\sim\!23\times$ & Tab.~\ref{tab:trainable_c_comparison} \\
\bottomrule
\end{tabular}

\smallskip
{\footnotesize \raggedright
$^*$Data-driven setting (no hard-constrained initial conditions); all
other rows are forward problems.
$^{\dagger}$Metric for this row is maximum position error, not RMSE.
Percentile RAR is compared against standard RAR; it improves
generalization rather than best-case RMSE
(Table~\ref{tab:lagrange_rar_comparison}).\par}
\end{table}

% ============================================================
\section{\texorpdfstring{Orbit discovery from sparse, noisy data}{Orbit discovery from sparse, noisy data}}
\label{sec:orbit_discovery}
% ============================================================

The forward experiments in Section~\ref{sec:experiments} established that PINNs
can accurately approximate known periodic orbits when initial conditions are
provided. We now turn to our central question: what happens when initial conditions
are \emph{removed} and the network must discover a dynamical regime from sparse,
noisy data alone?

\paragraph{The under-determined regime.}
Removing the initial conditions changes the character of the problem. With
hard-constrained initial conditions~\eqref{eq:hard_constraint}, the solution
is unique and the network need only approximate it; the optimization has a
single target. Without them, 90 noisy samples at $20\%$ noise do not
determine a trajectory---infinitely many smooth curves pass within the
noise band---so the network must select one, and the selection is made by
the ODE residual rather than by the data. Two consequences follow. First,
the residual is doing work that the data cannot: a purely data-driven
network has no mechanism for selecting a physical trajectory at all.
Appendix~\ref{app:pinn_vs_nn} illustrates this in the easier
hard-constrained setting, where an unconstrained network already violates
energy conservation by two orders of magnitude more than a PINN on the
same samples; without initial conditions the unconstrained problem is
worse still. Second, the residual admits \emph{many} valid selections,
because the three-body problem itself has many periodic solutions
consistent with sparse observations---so the optimizer's choice among them
carries physical meaning.

A separate difficulty is one of scale. The residual gradients dominate
the data gradients early in training, when the network output is far from
any trajectory, which is why the trainable scaling parameter of
Section~\ref{subsec:trainable_C} is needed for the data term to have
any effect.

Throughout this section we say an orbit is \emph{discovered} when the
PINN-inferred state, used as an initial guess, refines to a periodic
solution with closure error $\delta_T < 10^{-8}$ \emph{and} the refined
state is identifiable as a member of a known family via its
scale-invariant period. Here $\delta_T = \lVert \vect{z}(T) -
\vect{z}(0)\rVert_2$ is the closure error, the distance between the
initial and final states after one period, and the \emph{refinement
distance} is the relative $\ell_2$ distance between the PINN-inferred and
refined initial conditions---a measure of how much of the work the
refinement had to do. Runs falling in the ``other'' category of the tables
below close to $\delta_T < 10^{-8}$ but are not identifiable against any
catalogued family, and we do not count them as discoveries; this is a
statement about identification, not about periodicity.

\subsection{Orbit discovery without initial conditions}
\label{subsec:orbit_discovery_main}

When we remove the initial condition constraints and train a PINN with only the
gravitational ODE residual and the same noisy data---using a
$3 \times 64$ network, AdamW
($\eta = 10^{-3}$, weight decay $= 0.004$), loss weights $w_r = 0.1$,
$w_d = 40$, $\varepsilon = 10^{-9}$, and output biases
$\vect{b}_\text{out} = (-0.2, 0.2, 0, 0, 0.2, -0.2)$---a consistent pattern
emerges. The network must now simultaneously discover \emph{which} orbit to
follow and \emph{how} to follow it. We observe that this under-determined problem
is resolved differently depending on the random initialization seed, as
illustrated here with one representative run per distribution:

\begin{itemize}
    \item \textbf{Glorot Uniform initialization (single run):} The PINN places the three bodies
    on a line ($d_{12} + d_{23} \approx d_{13}$ to two decimal places) and produces
    Euler-type collinear orbits with body 3 approximately stationary at the center.

    \item \textbf{Glorot Normal initialization (single run):} The PINN places the three bodies
    at the vertices of an approximate equilateral triangle
    ($d_{12} \approx d_{13} \approx d_{23}$ to one decimal place) and produces
    Lagrange-type orbits.
\end{itemize}

\begin{figure}[H]
    \centering
    \begin{subfigure}[b]{0.48\textwidth}
        \centering
        \includegraphics[width=\textwidth]{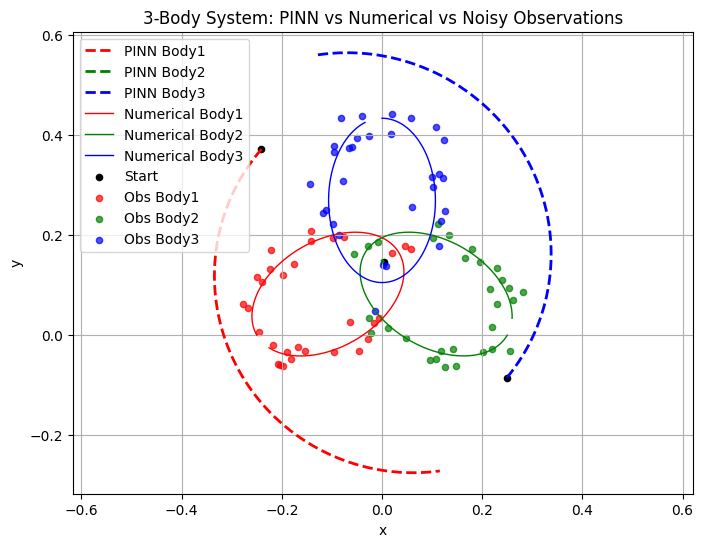}
        \caption{A Glorot Uniform run: Euler orbits}
    \end{subfigure}
    \hfill
    \begin{subfigure}[b]{0.48\textwidth}
        \centering
        \includegraphics[width=\textwidth]{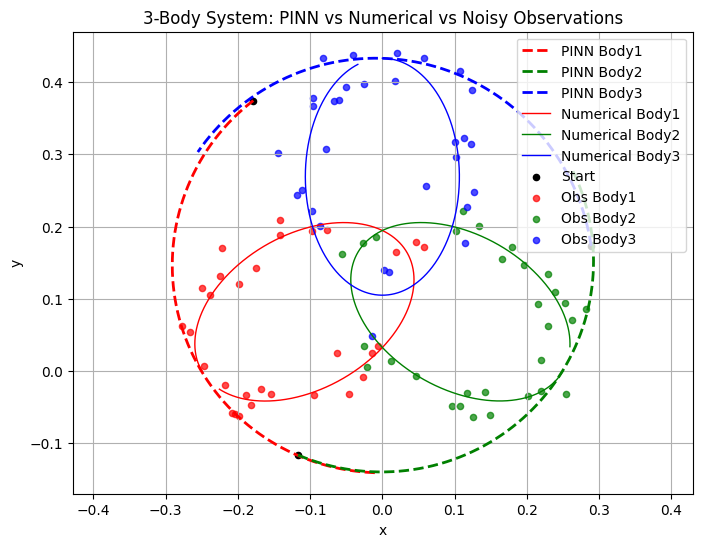}
        \caption{A Glorot Normal run: Lagrange orbits}
    \end{subfigure}
    \caption{Autonomous orbit discovery without initial conditions. Trained on the
    same noisy data (dots) from Lagrange orbits with identical hyperparameters but
    different weight initialization: (a) one Glorot Uniform run yielding
    Euler collinear orbits with $d_{12} + d_{23} \approx d_{13}$;
    (b) one Glorot Normal run yielding Lagrange equilateral orbits with
    $d_{12} \approx d_{13} \approx d_{23}$. These are single runs and do
    not indicate a systematic init--family mapping: the 200-seed study
    (Section~\ref{subsec:statistical_robustness}) finds no significant
    association ($p = 0.620$). Dashed lines: PINN predictions;
    solid lines: numerical reference.}
    \label{fig:orbit_discovery}
\end{figure}

Table~\ref{tab:orbit_discovery_metrics} reports the quantitative metrics
for these two runs and Figure~\ref{fig:orbit_discovery} shows the resulting
trajectories.
The data, network architecture, optimizer, learning rate, and all
other hyperparameters are \emph{identical} between the two experiments---only the
distribution from which the initial weights are drawn differs.
These are single runs, and we do not read a systematic
distribution-to-family mapping into them; the seed ensembles of
Section~\ref{subsec:statistical_robustness} address that question
directly, under a different loss weighting and learning rate.

That two runs differing only in their random seed land on different
orbit families is what one would expect if the objective were multi-modal.
The composite loss
$\mathcal{L}(\vect{\theta}) = w_r \mathcal{L}_r(\vect{\theta}) + w_d \mathcal{L}_d(\vect{\theta})$
admits many near-optimal configurations, because many periodic solutions
of the three-body problem are compatible with 90 noisy samples; the seed
selects among them. In standard supervised learning, different seeds
converge to functionally equivalent minima---the network computes
essentially the same function either way. Here the minima correspond to
qualitatively different physical solutions, so the choice is visible in
the output as a change of orbit family rather than as a change in
accuracy. We do not claim a causal mechanism.

\begin{table}[H]
\centering
\caption{Orbit discovery metrics: quantitative comparison of PINN solutions
obtained with different weight initialization distributions. Data: 90 noisy
observations (20\% noise) from Lagrange orbits. No initial conditions provided.
Distances $d_{ij}$ are inter-body separations at $t = 0$ (inferred by the PINN).
The collinearity measure is
$|d_{\min} + d_{\mathrm{mid}} - d_{\max}|$, where
$d_{\min} \leq d_{\mathrm{mid}} \leq d_{\max}$ are the sorted inter-body
distances; values near zero
indicate collinear configurations. The equilateral measure is
$\max|d_{ij} - \bar{d}|/\bar{d}$; values near zero indicate equilateral
configurations.}
\label{tab:orbit_discovery_metrics}
\begin{tabular}{lcc}
\toprule
\textbf{Metric} & \textbf{Glorot Uniform} & \textbf{Glorot Normal} \\
\midrule
Discovered orbit family & Euler (collinear) & Lagrange (equilateral) \\
Final total loss & $2.19 \times 10^{-1}$ & $1.25 \times 10^{-1}$ \\
Data loss $\mathcal{L}_d$ & $2.14 \times 10^{-1}$ & $1.22 \times 10^{-1}$ \\
ODE residual loss $\mathcal{L}_r$ & $4.50 \times 10^{-3}$ & $3.54 \times 10^{-3}$ \\
Collinearity measure & $1.16 \times 10^{-4}$ & $\sim 10^{0}$\;(0.5) \\
Equilateral measure & $\sim 10^{0}$\;(0.5) & $1.03 \times 10^{-2}$ \\
$d_{12}$ at $t = 0$ & $0.345$ & $0.498$ \\
$d_{13}$ at $t = 0$ & $0.678$ & $0.491$ \\
$d_{23}$ at $t = 0$ & $0.334$ & $0.500$ \\
\bottomrule
\end{tabular}
\end{table}

One caveat applies to all the experiments that follow. The
cross-family BHH discovery (Section~\ref{subsec:bhh_discovery}) uses
different loss weights and still produces valid periodic orbits, which
indicates the phenomenon is not confined to a single configuration; but
we varied only loss weights, activation and initialization, so we cannot
separate the geometry of the objective from the optimizer's trajectory
through it.
\subsection{Seed ensembles under two initialization distributions}
\label{subsec:statistical_robustness}

To validate the multi-basin structure observed in the single-run
experiments of Section~\ref{subsec:orbit_discovery_main}, we repeat the orbit
discovery experiment over $N = 100$ independent random seeds under two
initialization distributions---Glorot Uniform (GU) and Glorot Normal
(GN)---both using noisy Lagrange training data, with the following
setup: $3 \times 64$ network with $\sin$ activation, AdamW optimizer
($\eta = 10^{-4}$, weight decay $= 0.004$), equal loss weights
$w_r = w_d = 1$, softening $\varepsilon = 10^{-9}$, and
$4 \times 10^4$ epochs (with an additional $2 \times 10^4$ epochs at
$\eta = 10^{-5}$ for seeds that required further convergence).
Note that this setup differs from the single-run experiments of
Section~\ref{subsec:orbit_discovery_main}, which used $w_r = 0.1$,
$w_d = 40$ and $\eta = 10^{-3}$; the equal-weight, lower-learning-rate
configuration was adopted for the large-scale seed study.
Each discovered orbit is classified according to two geometric
criteria, evaluated at 5000 points along the \emph{least-squares refined}
trajectory rather than on the raw network output. This distinction matters
for reading the thresholds below: the refined trajectory is a numerically
integrated periodic solution, so its geometric residuals reach machine
precision, whereas the corresponding measures on the PINN output are of
order $10^{-3}$ (cf.\ Table~\ref{tab:orbit_discovery_metrics}). The
criteria are:
(i)~Euler-type (collinear), if the collinearity measure
$|d_{\min} + d_{\mathrm{mid}} - d_{\max}|$ is $\mathcal{O}(10^{-16})$ (i.e., machine
precision); (ii)~Lagrange-type (equilateral), if the equilateral deviation
$\max|d_{ij} - \bar{d}|/\bar{d}$ is $\mathcal{O}(10^{-11})$. Orbits
satisfying neither criterion are further identified via the scale-invariant
period $T^* = T|E|^{3/2}$ and angular momentum
$L^* = L_z |E|^{1/2}$ on the periodic orbit invariant map
(Figure~\ref{fig:orbit_invariant_map}; see also Appendix~\ref{app:gn_results}).
Orbits whose $T^*$ values match known Broucke--Hadjidemetriou--H\'{e}non
family members are classified as BHH-like; those matching the Li--Liao
catalog \citep{Li2017catalog}---class~I.A.1 being the figure-eight
choreography---are identified accordingly; the remainder as
``other.'' We retain the ``-like'' qualifier for the BHH classifications:
$T^*$ is a single scalar and does not by itself determine an orbit, so
while the geometric and $T^*$-based evidence is strong, independent
confirmation by continuation has not been performed. The figure-eight
identification is on firmer ground---its $T^*$ and $L^*$ follow in closed
form and are reproduced to seven significant digits
(Section~\ref{subsec:liliao_discovery})---as are the Euler and Lagrange
classifications, whose $T^*$ values are likewise exact
(Eq.~\ref{eq:Tstar_theory}).

\subsubsection{Glorot Uniform initialization (100 seeds)}

Table~\ref{tab:seed_results_gu} summarizes the GU results.
The majority of seeds ($71\%$) converge to
Lagrange-type orbits, consistent with the training data. However, $11\%$ of
seeds produce Euler collinear orbits \emph{despite} the Lagrange training
data, and $12\%$ discover BHH-like orbits---a topologically distinct family not present
in the training data---indicating that cross-family convergence recurs across seeds rather than
occurring in isolation.

\begin{table}[H]
\centering
\caption{Orbit family distribution over $N = 100$ independent seeds
with Glorot Uniform initialization and noisy Lagrange training data. Mean
training loss, distinct scale-invariant periods $T^*$, and mean refinement
distance (relative $\ell_2$ norm between PINN-inferred and least-squares
refined initial conditions) are reported per family. All refined orbits
achieve closure errors $\delta_T < 10^{-8}$.}
\label{tab:seed_results_gu}
\footnotesize
\begin{tabular}{lrrcp{4.8cm}r}
\toprule
\textbf{Family} & \textbf{Count} & \textbf{\%} & \textbf{Mean loss}
& \textbf{Distinct $T^*$ values} & \textbf{Refine.\ (\%)} \\
\midrule
Lagrange & 71 & 71 & 0.274 & 6.664 & 7.30 \\
Euler    & 11 & 11 & 0.483 & 7.854 & 2.70 \\
BHH-like & 12 & 12 & 0.395 & 7.327,\; 9.651,\; 11.817,\; 19.302 & 22.67 \\
Other    & 6 & 6 & 0.474 & 4.960,\; 7.218,\; 9.352,\; 9.686,\; 9.918 & 57.93 \\
\bottomrule
\end{tabular}
\end{table}

The Euler and Lagrange orbits admit an exact reduction that makes
these invariants easy to interpret. Both arise from \emph{central
configurations}, and the corresponding motions are \emph{homographic}: the
configuration retains its shape while its overall scale and orientation
follow a two-dimensional Kepler orbit
\citep{MeyerHallOffin2009, Moeckel2014cc}. For three equal unit masses the
reduction gives an effective Kepler parameter $\mu = 3^{-1/2}$ in the
Lagrange case and $\mu = 5/4$ in the Euler case, from which
\begin{equation}
    T^*_{\text{Lag}} = \frac{6\pi}{2^{3/2}} = 6.664324,
    \qquad
    T^*_{\text{Eul}} = \frac{5\pi}{2} = 7.853982 .
    \label{eq:Tstar_theory}
\end{equation}
Because $T$ and $E$ both depend only on the semi-major axis of the
underlying Kepler orbit, $T^*$ is independent of the eccentricity as well
as of the physical scale. Recovering the values
in~\eqref{eq:Tstar_theory} therefore identifies the \emph{family} but
carries no information about which member was found; the reported values
agree with~\eqref{eq:Tstar_theory} to eight significant digits.
Convergence quality is instead measured by the refinement distance
($7.30\%$ for Lagrange, $2.70\%$ for Euler, the latter reflecting the
simplicity of collinear configurations).

\begin{figure}[H]
    \centering
    \includegraphics[width=0.55\textwidth]{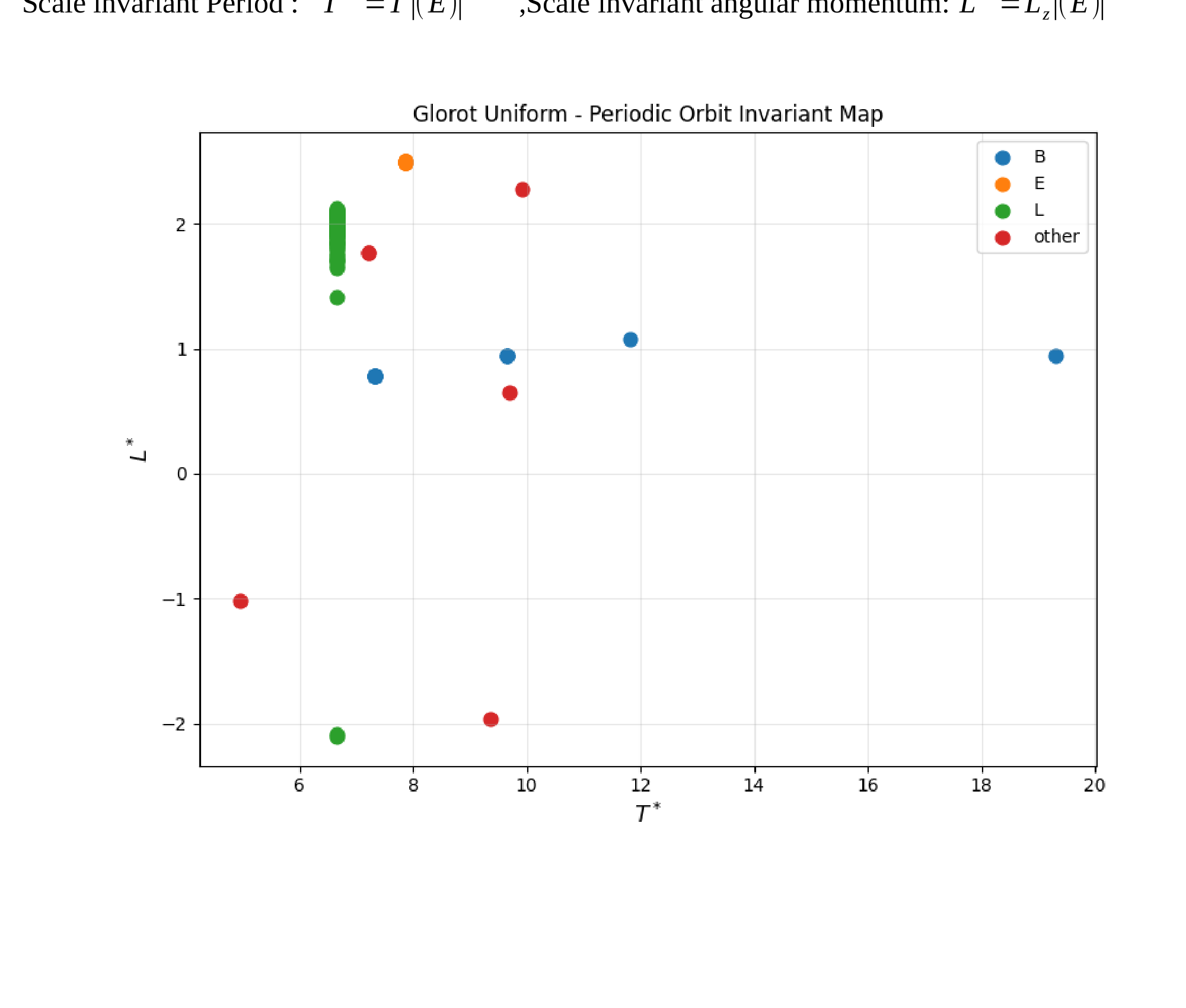}
    \caption{Glorot Uniform: periodic orbit invariant map for all 100
    seeds, plotted with \emph{signed} $L^*$ (the companion BHH-data map,
    Figure~\ref{fig:orbit_invariant_map_bhh}, uses $|L^*|$).
    Lagrange orbits (green) share $T^* = 6.664324$ but spread vertically
    over $|L^*| = 1.41$--$2.12$: since $L^* = (3/\sqrt{2})\sqrt{1-e^2}$,
    this axis resolves the eccentricity of the recovered member
    ($e = 0.03$--$0.75$), which $T^*$ cannot. One seed appears at
    $L^* \approx -2.1$, a retrograde solution. Euler (orange) sits at
    $(7.85, 2.49)$; BHH-like (blue) occupy four distinct $T^*$ values at
    $|L^*| \approx 0.8$--$1.1$; ``other'' (red) are isolated.}
    \label{fig:orbit_invariant_map}
\end{figure}

The member within a family is resolved by $L^*$. Since
$L_z \propto \sqrt{\mu\,a(1-e^2)}$ and $|E|^{1/2} \propto a^{-1/2}$, the
same reduction gives
$L^* = (3/\sqrt{2})\sqrt{1-e^2}$ for Lagrange and
$L^* = (5/2)\sqrt{1-e^2}$ for Euler, where $e$ is the eccentricity of the
underlying Kepler orbit---a quantity that is well defined for homographic
motions, though not for the three-body problem in general. This makes the
vertical axis of Figure~\ref{fig:orbit_invariant_map} an eccentricity
axis, and the spread along it is informative: the 71 Lagrange seeds
occupy $|L^*| = 1.41$--$2.12$, i.e.\ $e = 0.03$--$0.75$, whereas the
training orbit itself has $e = 0.76$ ($|L^*| = 1.379$). The PINN does not
merely reproduce the observed member; it recovers a broad section of the
one-parameter Lagrange family, biased toward orbits \emph{less} eccentric
than the data. The Euler seeds, by contrast, cluster tightly at
$|L^*| = 2.485$--$2.499$, within $e \leq 0.11$ of the circular
configuration.
The 12 BHH-like seeds span
\emph{four} distinct invariant periods ($T^* \approx 7.327$, $9.651$,
$11.817$, $19.302$), indicating that the PINN accesses multiple BHH-like
sub-families from identical training data. The ``other'' category (6~seeds)
exhibits a high mean training loss ($0.474$, second only to Euler's
$0.483$) and the largest
refinement distance ($57.93\%$). These orbits do close
($\delta_T < 10^{-8}$ after refinement), so they are genuine periodic
solutions; but the PINN-inferred state is so far from the refined one that
the network cannot be said to have located them---the refinement did most
of the work. We therefore do not count them as discoveries.

% FIGURE: GU Orbit family heat map across seeds
\begin{figure}[H]
    \centering
    \includegraphics[width=0.85\textwidth]{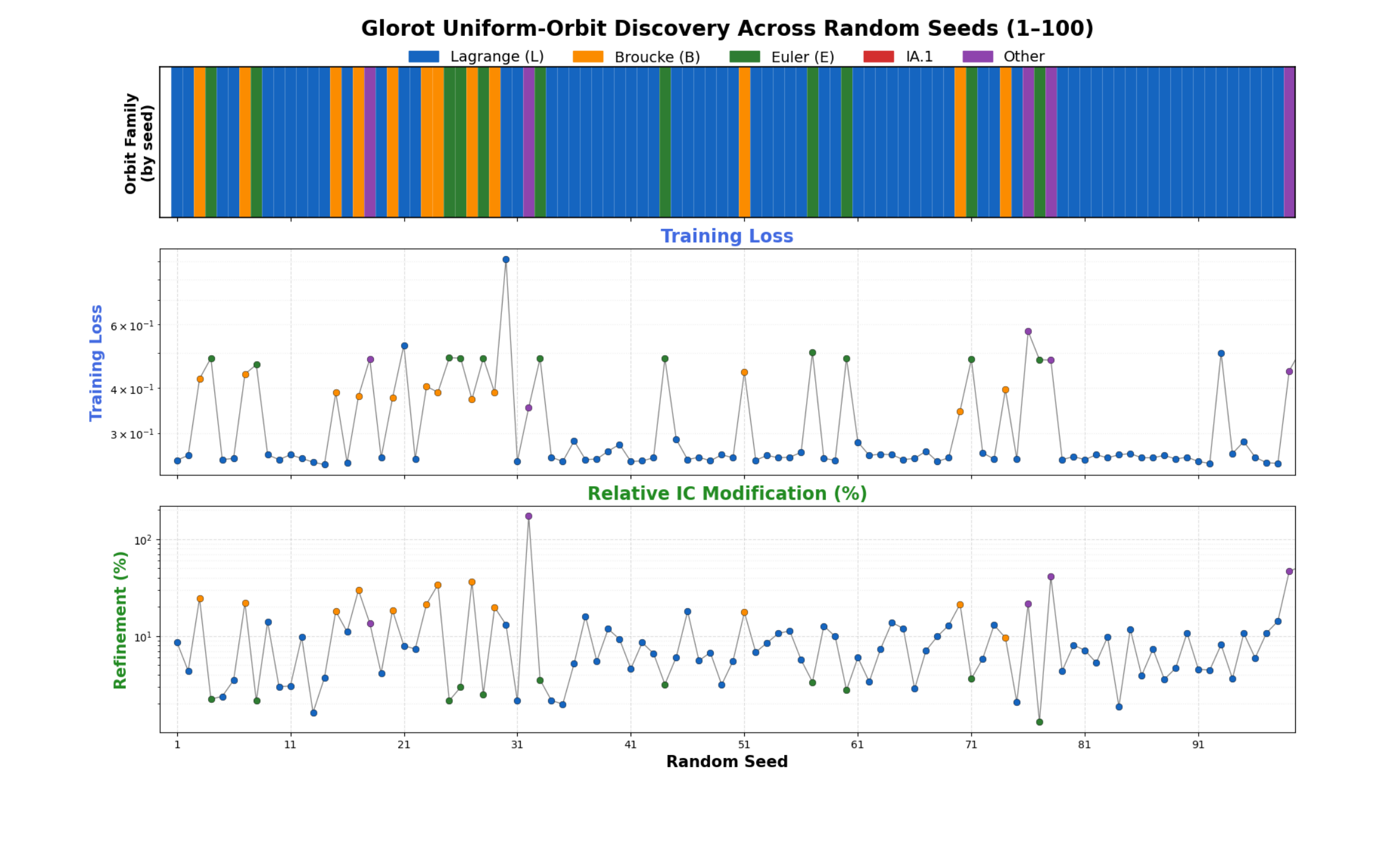}
    \caption{Glorot Uniform: orbit family distribution across 100 random seeds
    (top: color-coded bar), with per-seed training loss (middle) and
    refinement distance (bottom). Non-Lagrange seeds exhibit systematically higher
    training loss.}
    \label{fig:seed_heatmap}
\end{figure}

% FIGURE: GU Periodic orbit invariant map (T* vs L*)

\subsubsection{Glorot Normal initialization (100 seeds)}

Repeating the experiment with Glorot Normal initialization yields a
similar distribution: Lagrange $65\%$, BHH-like $13\%$,
Euler $11\%$, other $10\%$, plus a single figure-eight orbit (seed~66;
Section~\ref{subsec:liliao_discovery}). Full GN results---orbit family
table (Table~\ref{tab:seed_results_gn}), seed heatmap
(Figure~\ref{fig:seed_heatmap_gn}), and invariant map
(Figure~\ref{fig:orbit_invariant_map_gn})---are provided in
Appendix~\ref{app:gn_results}.

\subsubsection{Chi-squared test: initialization distribution does not
affect orbit family selection}

The similarity between the GU and GN distributions invites a formal
statistical comparison. We construct a $4 \times 2$ contingency table by
merging the figure-eight orbit into the ``other'' category (to satisfy the
minimum expected frequency requirement) and perform a Pearson $\chi^2$ test
of independence (Table~\ref{tab:chi_squared}).

\begin{table}[H]
\centering
\caption{Chi-squared test of independence between initialization
distribution and orbit family. The figure-eight orbit (1~seed, GN) is merged
into the ``other'' category. The null hypothesis $H_0$ is that the
initialization distribution and orbit family are independent.}
\label{tab:chi_squared}
\begin{tabular}{lcccc}
\toprule
& \textbf{BHH-like} & \textbf{Euler} & \textbf{Lagrange} & \textbf{Other} \\
\midrule
\emph{Observed counts} \\
\quad Glorot Uniform & 12 & 11 & 71 & 6 \\
\quad Glorot Normal  & 13 & 11 & 65 & 11 \\
\midrule
\emph{Expected counts} \\
\quad (under $H_0$) & 12.5 & 11.0 & 68.0 & 8.5 \\
\midrule
\multicolumn{5}{l}{$\chi^2(3) = 1.775$, \quad $p = 0.620$, \quad Cram\'{e}r's $V = 0.094$} \\
\multicolumn{5}{l}{Minimum expected frequency $= 8.5$ \quad ($> 5$; assumptions satisfied)} \\
\bottomrule
\end{tabular}
\end{table}

The test yields $\chi^2(3) = 1.775$ with $p = 0.620$, well above the
conventional $\alpha = 0.05$ threshold. Cram\'{e}r's $V = 0.094$ indicates
negligible effect size. All standardized residuals are small
(maximum $|r| = 0.857$ for the ``other'' category). We therefore
\emph{fail to reject} $H_0$: the choice of Glorot Uniform versus Glorot
Normal initialization does not significantly affect the distribution of
orbit families recovered by the PINN. The observed differences between the
two experiments are consistent with random sampling variability.

This null result narrows the field of candidate explanations: whatever
governs the aggregate family frequencies, it is not the choice between
these two initialization distributions. It does not by itself identify
what does; Section~\ref{subsec:training_data_effect} supplies that
evidence directly. While the
single-run experiments of Section~\ref{subsec:orbit_discovery_main} showed
that \emph{different} seeds can lead to \emph{different} orbit families
(confirming multi-basin structure), the 200-seed validation finds no
significant difference in \emph{aggregate} basin selection frequencies
between Glorot Uniform and Glorot Normal. The training data exerts the
dominant influence, steering $65$--$71\%$ of seeds toward the Lagrange
family, while $23$--$25\%$ of seeds discover orbits from families
\emph{not present in the training data} (Euler, BHH-like, and figure-eight;
excluding the ``other'' category, which reflects incomplete convergence). Section~\ref{subsec:training_data_effect} tests this
hypothesis directly by changing the training data source.

\subsection{Training data significantly affects basin selection}
\label{subsec:training_data_effect}

To determine whether the training data---rather than the
initialization---shapes the basin structure, we repeat the 100-seed GU
experiment with noisy BHH training data (instead of Lagrange).
Table~\ref{tab:seed_results_bhh} summarizes the results.

\begin{table}[H]
\centering
\caption{Orbit family distribution over $N = 100$ independent seeds
with Glorot Uniform initialization and noisy BHH training data. Compare
with Table~\ref{tab:seed_results_gu} (Lagrange training data). As in
Table~\ref{tab:chi_squared}, the single figure-eight seed
($T^* = 9.238$) is merged into the ``other'' category.}
\label{tab:seed_results_bhh}
\footnotesize
\begin{tabular}{lrrcp{5.0cm}r}
\toprule
\textbf{Family} & \textbf{Count} & \textbf{\%} & \textbf{Mean loss}
& \textbf{Distinct $T^*$ values} & \textbf{Refine.\ (\%)} \\
\midrule
Lagrange  & 38 & 38 & 0.290 & 6.664 & 8.03 \\
Euler     & 17 & 17 & 0.372 & 7.854 & 1.79 \\
BHH-like  & 28 & 28 & 0.495 & 7.327,\; 9.651,\; 11.817,\; 14.652,\; 15.888,\; 17.847,\; 19.769 & 14.30 \\
Other     & 17 & 17 & 0.429 & 4.960,\; 7.218,\; 7.312,\; 9.238,\; 9.267,\; 9.352,\; 9.878,\; 11.781,\; 12.069,\; 12.097,\; 14.652 & 56.83 \\
\bottomrule
\end{tabular}
\end{table}

Switching from Lagrange to BHH training data
shifts the distribution: Lagrange orbits drop from $71\%$ to $38\%$,
while BHH-like orbits more than double from $12\%$ to $28\%$.
Euler orbits increase from $11\%$ to $17\%$ and ``other'' from $6\%$
to $17\%$. A $\chi^2$ test of independence
(Table~\ref{tab:chi_squared_data}) confirms that this shift is
statistically significant.

\begin{table}[H]
\centering
\caption{Chi-squared test of independence between training data
source and orbit family (both experiments use Glorot Uniform,
$N = 100$ seeds each). figure-eight seeds are merged into the ``other''
category in both rows. $H_0$: the training data source and orbit
family are independent.}
\label{tab:chi_squared_data}
\begin{tabular}{lcccc}
\toprule
& \textbf{BHH-like} & \textbf{Euler} & \textbf{Lagrange} & \textbf{Other} \\
\midrule
\emph{Observed counts} \\
\quad Lagrange data & 12 & 11 & 71 & 6 \\
\quad BHH data      & 28 & 17 & 38 & 17 \\
\midrule
\emph{Expected counts} \\
\quad (under $H_0$) & 20.0 & 14.0 & 54.5 & 11.5 \\
\midrule
\multicolumn{5}{l}{$\chi^2(3) = 22.94$, \quad $p < 0.001$, \quad Cram\'{e}r's $V = 0.339$} \\
\multicolumn{5}{l}{Minimum expected frequency $= 11.5$ \quad ($> 5$; assumptions satisfied)} \\
\bottomrule
\end{tabular}
\end{table}

The test yields $\chi^2(3) = 22.94$ with $p < 0.001$ and
Cram\'{e}r's $V = 0.339$ (moderate effect size). The largest standardized
residuals are for Lagrange orbits ($|r| = 2.24$) and BHH-like orbits
($|r| = 1.79$): Lagrange training data produces significantly more
Lagrange orbits, while BHH training data produces significantly more
BHH-like orbits. Under BHH training data, $55\%$ of seeds converge to a
family other than the one supplying the data, against $23\%$ under
Lagrange data. Combined with the null result of the initialization test
($p = 0.620$), these findings indicate that \emph{the training data source
is associated with a significant shift in the distribution of recovered
families, whereas the choice between the two initialization distributions
tested is not}.

\begin{figure}[H]
    \centering
    \includegraphics[width=0.85\textwidth]{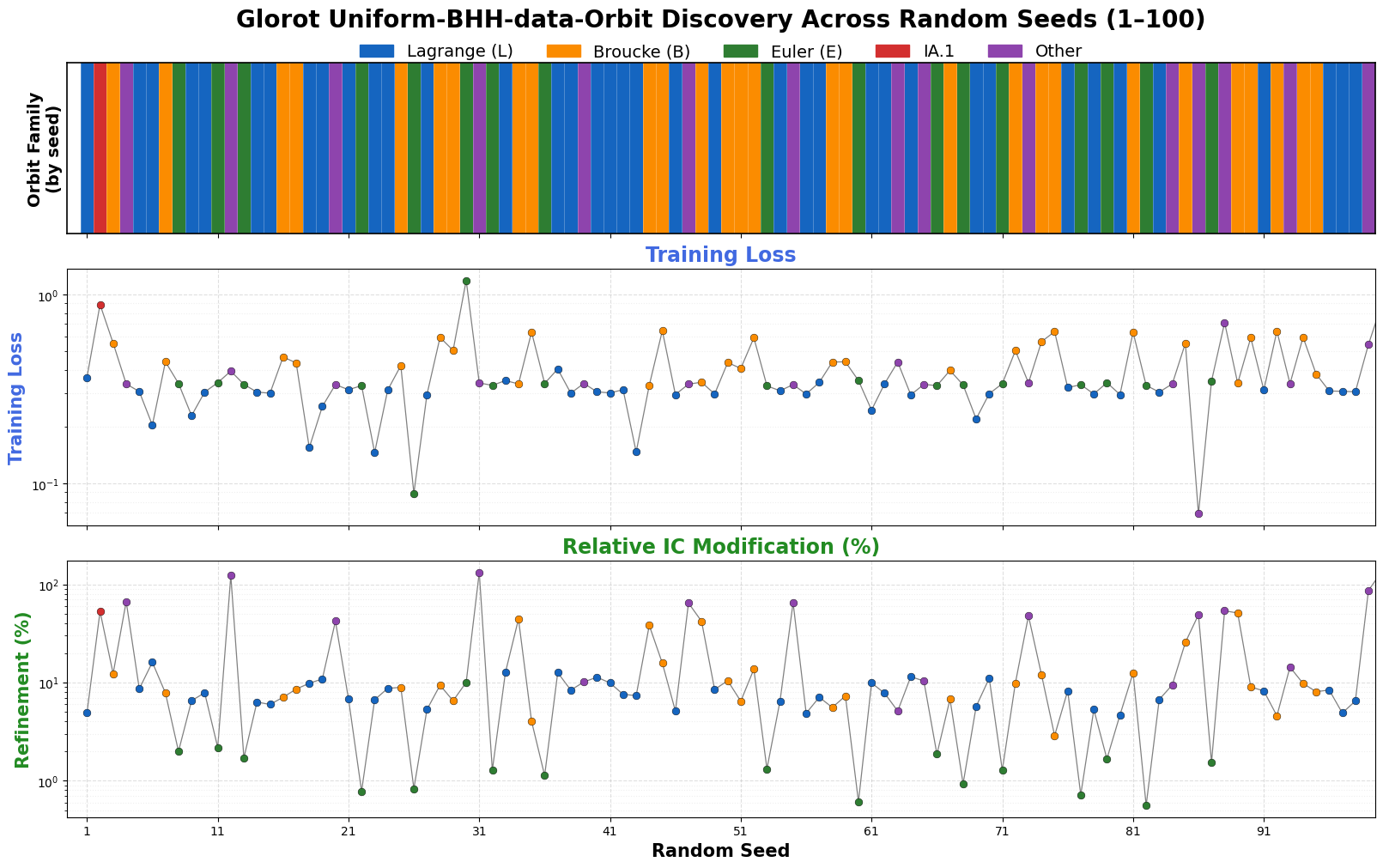}
    \caption{Glorot Uniform with BHH training data: orbit family
    distribution across 100 random seeds (top: color-coded bar), with
    per-seed training loss (middle) and refinement distance (bottom).
    Compare with Figure~\ref{fig:seed_heatmap} (Lagrange data):
    Broucke/BHH orbits (orange) are substantially more frequent.}
    \label{fig:seed_heatmap_bhh}
\end{figure}

\begin{figure}[H]
    \centering
    \includegraphics[width=0.55\textwidth]{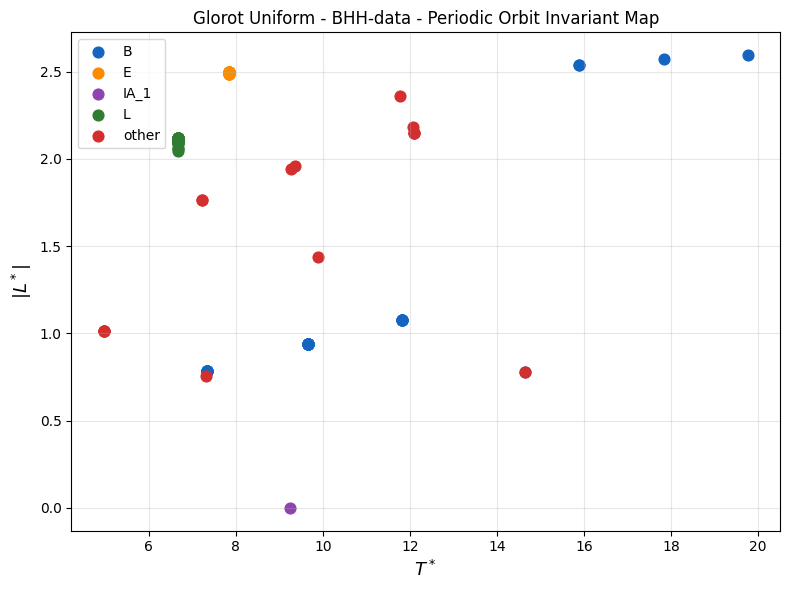}
    \caption{Glorot Uniform with BHH training data: periodic orbit invariant
    map ($T^*$ vs.\ $|L^*|$) for all 100 seeds. BHH-like orbits (B, blue)
    split into a low-$|L^*|$ group ($|L^*| \approx 0.8$--$1.1$,
    $T^* \approx 7.3$--$11.8$) and a high-$|L^*|$ group
    ($|L^*| \approx 2.5$--$2.6$, $T^* \approx 15.9$--$19.8$); the
    $T^* = 14.652$ seeds listed in Table~\ref{tab:seed_results_bhh} plot
    with the ``other'' points.
    Lagrange orbits (L, green) cluster tightly at
    $(T^*, |L^*|) \approx (6.66, 2.1)$ and Euler (E, orange) at
    $(7.85, 2.49)$; a single figure-eight orbit (purple) appears at
    $(9.24, 0)$. Compare with the Lagrange-data invariant map
    (Figure~\ref{fig:orbit_invariant_map}).}
    \label{fig:orbit_invariant_map_bhh}
\end{figure}

\begin{figure}[H]
    \centering
    \includegraphics[width=\textwidth]{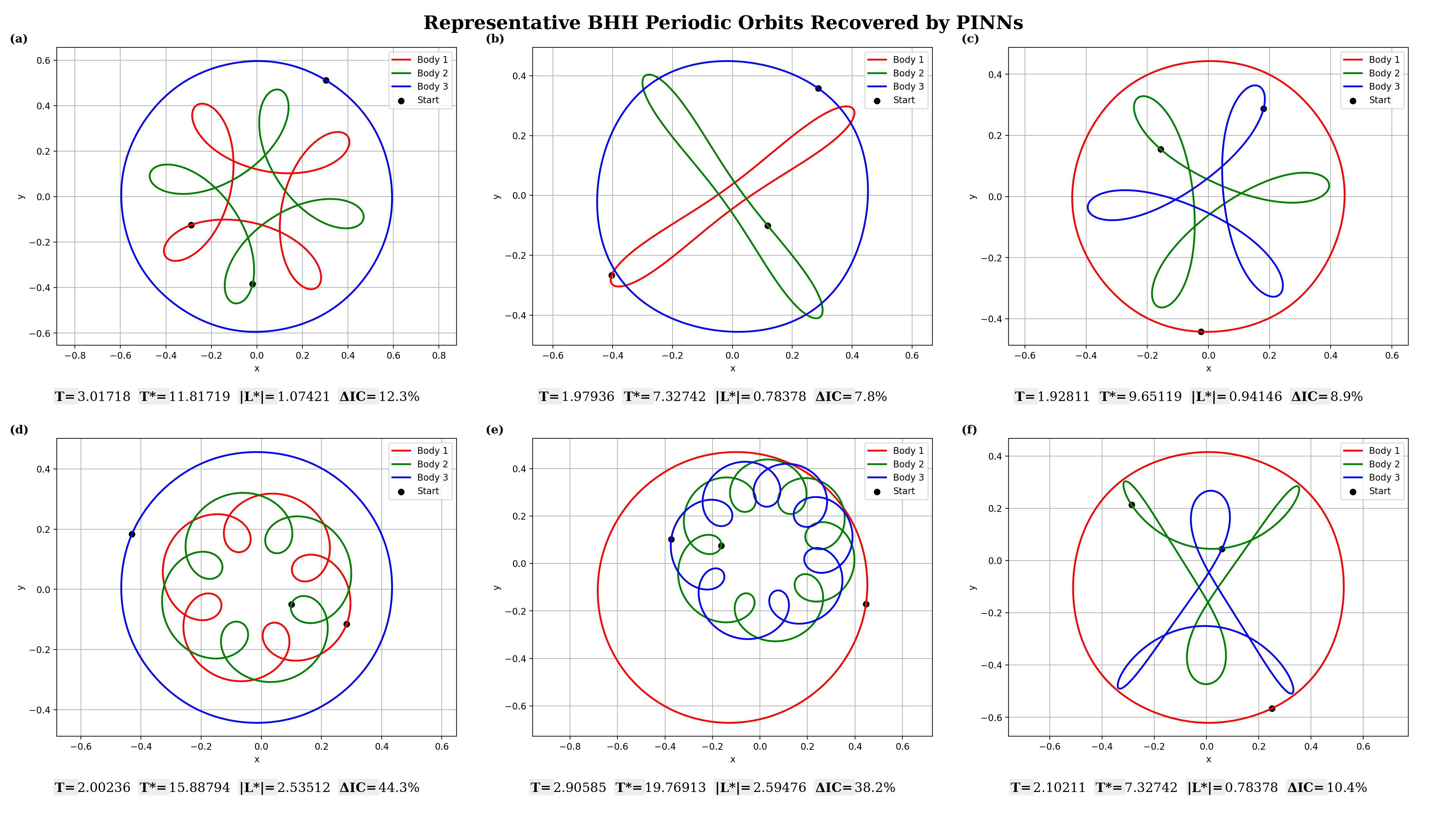}
    \caption{Representative BHH periodic orbits recovered by PINNs
    trained on noisy BHH data (Glorot Uniform). Each panel shows the
    least-squares refined trajectory with period $T$, scale-invariant
    period $T^*$, scale-invariant angular momentum $|L^*|$, and relative
    initial-condition refinement distance $\Delta\mathrm{IC}$.
    Panels~(a)--(c) and~(f) have $\Delta\mathrm{IC} < 13\%$, indicating
    close agreement between PINN-inferred and refined initial conditions;
    panels~(d) and~(e) ($\Delta\mathrm{IC} > 38\%$) represent orbits
    where substantial refinement was needed.}
    \label{fig:bhh_orbits_gallery}
\end{figure}

The BHH-data experiment also reveals a richer orbit taxonomy:
28 BHH-like seeds span \emph{seven} distinct invariant periods
(vs.\ four with Lagrange data; Table~\ref{tab:seed_results_gu}),
and the ``other'' category contributes 11 distinct $T^*$ values
(vs.\ five). The shift in orbit family distribution is visually apparent
in the seed heatmap (Figure~\ref{fig:seed_heatmap_bhh}), which contrasts
sharply with the Lagrange-data heatmap
(Figure~\ref{fig:seed_heatmap}), and in the invariant map
(Figure~\ref{fig:orbit_invariant_map_bhh}).
Figure~\ref{fig:bhh_orbits_gallery}
shows representative BHH orbits recovered from this experiment, spanning
a wide range of morphologies and invariant periods.

\subsection{Cross-family orbit discovery: figure-eight to BHH}
\label{subsec:bhh_discovery}

The seed studies above quantify how often cross-family convergence
occurs; we now examine one such case in detail, to establish that the
recovered orbit is a genuine periodic solution rather than a
classification artifact. We train a
$3 \times 64$ network with $\sin$ activation, Glorot Normal initialization,
and 90 noisy figure-eight observations (20\% noise; $w_r = 1$, $w_d = 10$,
no hard constraints or Fourier features). The PINN converges to a solution
resembling the Broucke--Hadjidemetriou--H\'{e}non (BHH) family
\citep{Broucke1975, Li2021BHH}---a topologically more complex orbit in which
one body traces a large outer loop while the other two execute smaller inner
loops.

\begin{figure}[H]
    \centering
    \begin{subfigure}[b]{0.48\textwidth}
        \centering
        \includegraphics[width=\textwidth]{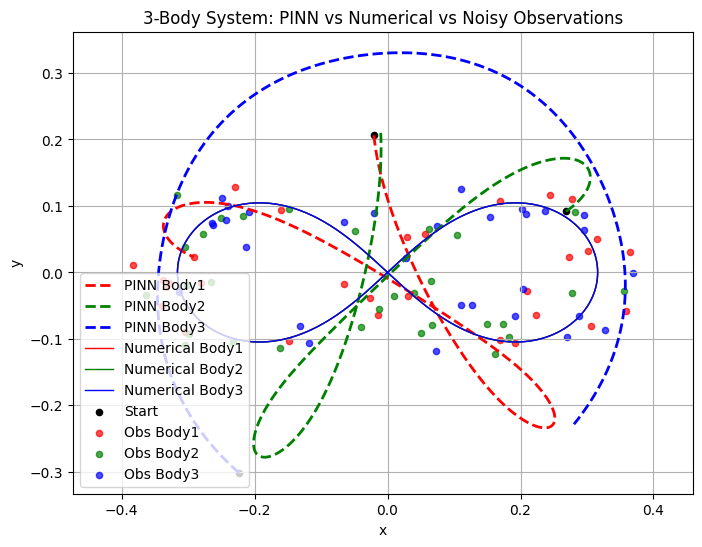}
        \caption{PINN solution vs.\ noisy figure-eight data}
    \end{subfigure}
    \hfill
    \begin{subfigure}[b]{0.48\textwidth}
        \centering
        \includegraphics[width=\textwidth]{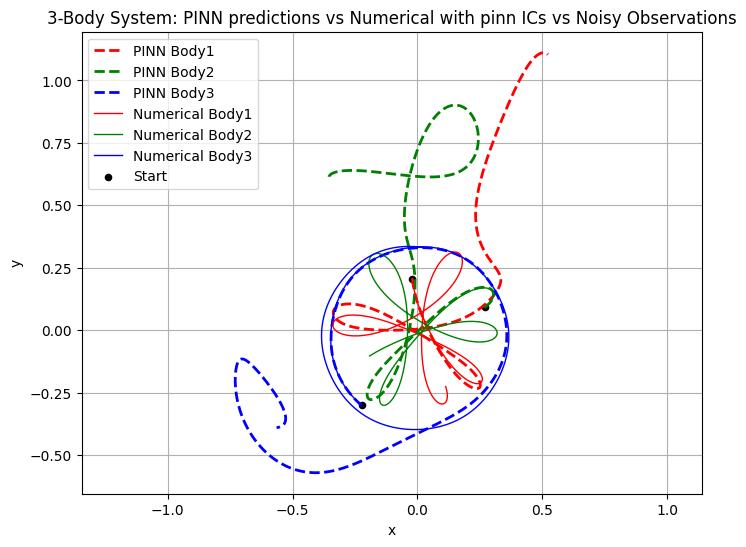}
        \caption{Numerical integration with PINN-discovered ICs}
    \end{subfigure}
    \caption{Cross-family orbit discovery. (a) PINN trained on 90
    noisy observations (dots) from the figure-eight orbit discovers a
    BHH-type orbit (dashed lines). (b) Forward integration with the
    least-squares refined initial conditions confirms BHH family
    topology (Table~\ref{tab:bhh_ics}).}
    \label{fig:bhh_discovery}
\end{figure}

Least-squares refinement of the PINN-inferred initial conditions
(Appendix~\ref{app:bhh_liliao_ics}, Table~\ref{tab:bhh_ics}) yields a
periodic orbit with $T \approx 1.4946$ and closure error
$\delta_T \approx 9.4 \times 10^{-10}$, confirming that the PINN-discovered
solution lies in the basin of attraction of a true BHH periodic orbit.
Forward integration with these initial conditions
(Figure~\ref{fig:bhh_discovery}) confirms the BHH topology. The PINN
discovers this orbit in the inertial frame; in a rotating frame
($\omega = -0.35$) the trajectory matches the canonical BHH topology.

The behaviour is not limited to a single configuration: repeating with
Glorot Uniform initialization on BHH data produces a more complex
four-lobed BHH variant, indicating that BHH-family convergence is not an
artifact of one particular run.

The 200-seed experiment (Section~\ref{subsec:statistical_robustness})
classifies 12 (GU) and 13 (GN) seeds as BHH-like.
Figure~\ref{fig:bhh_orbits_gallery} illustrates the morphological variety
of this family using the BHH-data ensemble
(Section~\ref{subsec:training_data_effect}), where the morphologies
range from simple near-symmetric configurations to complex multi-lobed
trajectories. This suggests that the ``BHH-like'' basin encompasses a
continuum of related periodic orbits rather than a single solution.

\subsection{Discovery of the figure-eight choreography from Lagrange data}
\label{subsec:liliao_discovery}

\begin{figure}[H]
    \centering
    \includegraphics[width=0.45\textwidth]{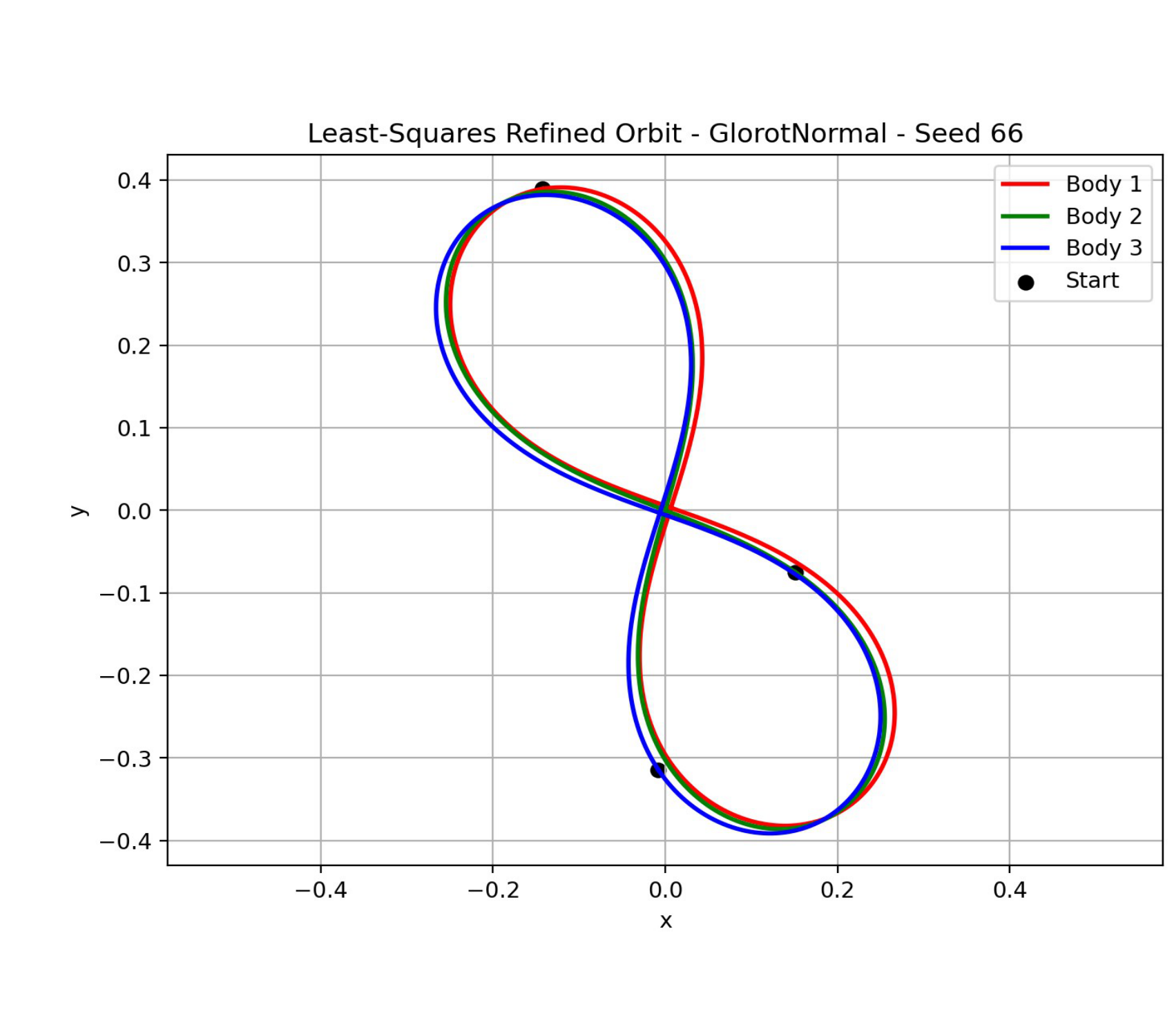}
    \caption{Least-squares refined orbit discovered by a PINN trained
    on noisy Lagrange data with Glorot Normal initialization
    (seed~66 out of 100). The two-lobed
    topology matches the first orbit in the Li--Liao catalog
    \citep{Li2017catalog}: $T^* = 9.237683$,
    $T = 1.518234$, $E = -3.332853$,
    $\delta_T = 1.07 \times 10^{-10}$.}
    \label{fig:liliao_orbit}
\end{figure}

During the seed ensembles of
Section~\ref{subsec:statistical_robustness}, a PINN trained on noisy
Lagrange data with Glorot Normal initialization (seed~66) converged to an
orbit topologically distinct from anything in its training data: a
two-lobed curve traced by all three bodies
(Figure~\ref{fig:liliao_orbit}).

The scale-invariant period $T^* = T|E|^{3/2}$ \citep{Li2017catalog}
identifies it. Least-squares refinement gives $T^* = 9.2376827$ with
$L^* = 0$, and the standard figure-eight choreography of
\citet{Moore1993} and \citet{Chenciner2000}---initial conditions
$\vect{r}_1 = (0.97000436, -0.24308753)$,
$\dot{\vect{r}}_3 = (-0.93240737, -0.86473146)$, $T = 6.3259140$---has
$T^* = 9.2376812$ and $L^* = 0$ exactly. The two agree to seven
significant digits, and the recovered period $T = 1.518234$ is
reproduced to six decimals by rescaling the reference orbit through
Eq.~\eqref{eq:scaling_law}. The discovered orbit is therefore the
figure-eight, which appears as class~I.A.1---the first entry---of the
Li--Liao catalog \citep{Li2017catalog, Liao2022}. Its closure error is
$\delta_T = 1.07 \times 10^{-10}$ (refined initial conditions in
Appendix~\ref{app:bhh_liliao_ics}, Table~\ref{tab:liliao_ics}).

This closes a loop in the paper. The figure-eight is the orbit whose
forward solution required the Fourier feature encoding of
Section~\ref{subsec:fourier}; here it is recovered in the inverse
direction, from noisy Lagrange observations, by a network given no
initial conditions and no knowledge of its period. Together with the
converse case of Section~\ref{subsec:bhh_discovery}---figure-eight data
yielding a BHH orbit---it shows that cross-family convergence is not
confined to one direction between two particular families.

Unlike the single-run BHH cross-family discovery
(Section~\ref{subsec:bhh_discovery}), this orbit was found within the
100-seed GN experiment (Section~\ref{subsec:statistical_robustness})
using the standard seed-study setup ($\sin$ activation, Glorot Normal,
Lagrange data), demonstrating that cross-family convergence occurs even
without specialized hyperparameter choices.

% ============================================================
\section{Discussion}
\label{sec:discussion}
% ============================================================

\subsection{Methodological insights}

Our finding that the second-order ODE formulation outperforms the first-order
reduction contrasts with \citet{Gladstone2025}, who report first-order
advantages on PDE boundary-value problems (e.g., Helmholtz, Navier--Stokes). We hypothesize that the difference is
loss balance: the first-order system introduces velocity-definition residuals
($\dot{x}_i - v_{xi} = 0$) that are simple identity constraints, whereas the
gravitational $1/r_{ij}^2$ accelerations produce residuals orders of magnitude
larger, creating a severely imbalanced 12-term loss. The second-order
formulation, with only 6 acceleration-type residuals of comparable magnitude,
avoids this imbalance---yielding a ${\sim}\,13$-fold improvement in
training loss and a ${\sim}\,160$-fold reduction in maximum position error
for the Euler orbit (Table~\ref{tab:euler_comparison}). This suggests
that the optimal ODE order is problem-dependent, but a controlled ablation
isolating the loss-balance effect from other factors (e.g., AD depth) is
needed to confirm this explanation.

Our Fourier feature encoding differs from standard spectral bias
mitigation \citep{Rahaman2019, Wang2021FF} in that the frequencies are
fixed by the known orbital period rather than sampled or learned. Supplying
the fundamental and second harmonic turns a problem where a plain-input
PINN fails to generalize into one with stable convergence
(Table~\ref{tab:figure_eight_comparison}). We note that the $T/3$ phase
offsets we included add no representational content
(Section~\ref{subsec:fourier}); the benefit comes from the harmonics. Similarly, our percentile-based
RAR (60th--95th) is simpler than the RAD and RAR-D methods of
\citet{Wu2023} and proved effective for Lagrange orbits
(Table~\ref{tab:lagrange_rar_comparison}), where standard RAR concentrated
points at singularity-adjacent outliers.

\subsection{Orbit discovery and what governs it}

Our orbit discovery result (Section~\ref{sec:orbit_discovery}) extends
the multi-solution PINN framework of \citet{Zou2025} from
reaction-diffusion and fluid-flow bifurcation branches to a fundamentally
different setting: an inverse problem with unknown initial conditions and
physically distinct orbit families rather than branches of a single
solution. A further difference concerns the role of initialization.
\citet{Zou2025} attribute solution multiplicity to random initialization;
in our setting the seed likewise selects the solution, but the sampling
distribution it is drawn from does not measurably shift the frequencies
with which the families appear. We attribute this to the presence of a
data term in the loss, absent from their purely residual-driven
formulation.

This suggests a methodology for orbit exploration: training ensembles
of PINNs on sparse observations and cataloging the orbit families that
emerge---without requiring the accurate initial-condition guesses on which
gradient-based search methods depend \citep{Suvakov2013}. The cross-family
discovery of Section~\ref{subsec:bhh_discovery} strengthens this further: a
PINN trained on figure-eight data converges to BHH-type orbits
\citep{Broucke1975}, suggesting that PINNs may serve as complementary
tools for generating candidate initial conditions alongside the
large-scale numerical catalogs of \citet{Liao2022} and \citet{Li2021BHH}.

The 20\% noise level may play a regularizing role, though this
hypothesis has not been tested via a controlled ablation; experiments with
noise exceeding 50\% did not produce meaningful orbits.

\subsection{PINNs as complementary tools}

PINNs are not replacements for established numerical integrators, which
remain superior for well-posed initial-value problems. Their niche is
inverse problems with sparse, noisy data and unknown initial conditions,
where they can discover qualitatively different solutions from the same
observations. Compared to the hybrid surrogate approach of
\citet{SazUlibarrena2024}, our method embeds the physics directly and
trades efficiency for exploratory capability.
Concretely, the cost is a training run per candidate orbit; the return
is a set of initial conditions that numerical refinement converges on
without a prior guess.

\subsection{Limitations and future work}

All experiments used modest resources (Intel i7, 8GB RAM; Google Colab
TPU v6e-1 for accelerated runs), limiting network sizes and ensemble scales.
Conservation laws were not exploited as auxiliary loss terms.
Only two initialization distributions (Glorot Uniform/Normal) and two
training data sources (Lagrange, BHH) were tested, on equal-mass planar
configurations; whether initialization can be \emph{designed} to bias
basin selection (e.g.\ He, orthogonal) and whether the training-data
effect generalizes to other orbit families both remain open. Incorporating
Hamiltonian or Lagrangian structure into the architecture is a natural
extension we have not attempted.

% ============================================================
\section{Conclusions}
\label{sec:conclusions}
% ============================================================

PINNs trained on sparse, noisy observations \emph{without} initial
conditions recover periodic orbits of the gravitational three-body problem,
including orbit families absent from the training data. Over 200 seeds,
$23$--$25\%$ of runs converge to such families, and the inferred states
refine to genuine periodic solutions---the figure-eight choreography recovered from Lagrange data, another a BHH orbit closing to
$\delta_T \approx 9.4 \times 10^{-10}$. The utility of the approach lies in supplying candidate initial
conditions in regions of the solution space where no accurate guess is
available to seed conventional continuation or gradient search.

\paragraph{What is responsible.}
Two factors must be distinguished. Within a single experimental
configuration, the random seed alone produces three distinct orbit families from a
single dataset (Table~\ref{tab:seed_results_gu}): the seed selects the
solution. What the $\chi^2$ tests then address is the aggregate
distribution over seeds. Changing the training data source shifts that
distribution significantly ($p < 0.001$, Cram\'{e}r's $V = 0.339$);
switching between Glorot Uniform and Glorot Normal does not
($p = 0.620$, $V = 0.094$). An ensemble seeking orbit diversity should
therefore be diversified over training data, since substituting one Glorot
variant for the other leaves the family frequencies unchanged within the
resolution of our sample.

\paragraph{What this paper does not establish.}
The evidence is statistical, not analytical. We do not prove that the loss
landscape contains multiple basins; we observe convergence frequencies
over finite ensembles and verify the resulting orbits numerically. The
null result on initialization is a failure to reject, not a demonstration
of absence: with 100 seeds per group, small effects would go undetected.
The scope limitations of Section~\ref{sec:discussion} apply throughout.

This extends \citet{Zou2025}'s multi-solution framework from
reaction-diffusion and fluid flow to celestial mechanics, where the
discovered solutions are fundamentally different orbit families rather than
bifurcation branches. Table~\ref{tab:summary} summarizes the key results
across all orbit families; the ablation summary
(Table~\ref{tab:ablation_summary}) quantifies each methodological
component's contribution.

\begin{table}[H]
\centering
\caption{Summary of experiments and key results across all orbit families
studied in this work. ``Cross-family'' is the percentage of seeds
converging to a family other than the one supplying the training data,
excluding the ``other'' category.}
\label{tab:summary}
\small
\begin{tabular}{llccc}
\toprule
\textbf{Orbit} & \textbf{Key technique} & \textbf{Net} &
\textbf{Loss} & \textbf{RMSE} \\
\midrule
Two-body & Validation & $3{\times}64$ &
$1.79 \times 10^{-8}$ & $2.12 \times 10^{-7}$ \\
Euler & 2nd-order form. & $3{\times}128$ &
$2.72 \times 10^{-5}$ & $2.22 \times 10^{-4}$ \\
Lagrange & Modified RAR & $3{\times}64$ &
$4.57 \times 10^{-2}$ & $1.05 \times 10^{-2}$ \\
Figure-eight & Fourier + phase & $5{\times}64$ &
$1.27 \times 10^{-5}$ & $1.53 \times 10^{-4}$ \\
\midrule
\multicolumn{5}{l}{\emph{Data-driven single runs (no ICs, 20\% noise)}} \\
\midrule
Euler (disc.) & No ICs, $w_r{=}0.1$, $w_d{=}40$ & $3{\times}64$ &
$2.19 \times 10^{-1}$ & --- \\
Lagrange (disc.) & No ICs, $w_r{=}0.1$, $w_d{=}40$ & $3{\times}64$ &
$1.25 \times 10^{-1}$ & --- \\
BHH (cross-fam.) & No ICs, $w_r{=}1$, $w_d{=}10$ & $3{\times}64$ &
$2.97 \times 10^{-1}$ & --- \\
\bottomrule
\end{tabular}

\medskip

\small
\begin{tabular}{llcc}
\multicolumn{4}{l}{\emph{Seed ensembles (no ICs, 20\% noise, $w_r = w_d = 1$)}} \\
\toprule
\textbf{Experiment} & \textbf{Init.} & \textbf{Cross-family} &
\textbf{Result} \\
\midrule
100 seeds, Lagrange data & GU & $23\%$ & baseline \\
100 seeds, Lagrange data & GN & $25\%$ & vs.\ GU: $p = 0.620$ \\
100 seeds, BHH data & GU & $55\%$ & vs.\ Lagrange data: $p < 0.001$ \\
\bottomrule
\end{tabular}

\end{table}

\noindent \textbf{Open problems.} Three questions follow directly.
(i)~Why does the training data shape basin selection frequencies while the
initialization distribution does not---is there a characterization of the
data-plus-residual loss that predicts the basin weights?
(ii)~Can the basin structure be probed directly with loss landscape tools
(mode connectivity, Hessian spectra) rather than inferred from convergence
statistics? (iii)~Does the effect survive in settings where the underlying
multi-stability is better understood---the Duffing oscillator would be a
natural first test---and does it extend to unequal masses and three
dimensions? 

% ============================================================
% Code availability
% ============================================================
\section*{Code and data availability}
The source code, trained models, and Jupyter notebooks to reproduce all experiments
in this paper are publicly available on GitHub at:
\begin{center}
\url{https://github.com/nkollias/Pinns-three-body-problem}
\end{center}
The repository includes Python scripts and Jupyter notebooks
(DeepXDE/TensorFlow \citep{Lu2021}), configuration files, numerical
reference solutions, and plotting scripts to reproduce all figures.

% ============================================================
% Acknowledgments
% ============================================================
\section*{Acknowledgments}
This work was conducted as part of a Master's thesis at the Hellenic Open University.
Computational experiments were performed on a personal computer (Intel i7, 8GB RAM)
and a Google Colab TPU v6e-1.

\section*{CRediT authorship contribution statement}
\textbf{Nikolaos Kollias:} Software, Investigation, Validation, Formal analysis,
Visualization, Writing -- original draft.
\textbf{Nikolaos Matzakos:} Conceptualization, Methodology, Formal analysis,
Supervision, Writing -- review \& editing.

\section*{Declaration of competing interest}
The authors declare that they have no known competing financial interests or
personal relationships that could have appeared to influence the work reported
in this paper.

% ============================================================
% References
% ============================================================
\bibliographystyle{plainnat}
\bibliography{references}

% ============================================================
% APPENDICES
% ============================================================
\appendix

\section*{Appendices}
\noindent The appendices contain material supporting, but not required
for, the main argument. Appendix~\ref{app:forward_details} gives the full
forward-problem experiments summarized in Section~\ref{sec:experiments}.
Appendix~\ref{app:pinn_vs_nn} compares a PINN with an unconstrained
network on the same noisy data. Appendix~\ref{app:trainable_c_results}
reports the trainable scaling parameter results.
Appendix~\ref{app:gn_results} gives the Glorot Normal seed study in full,
mirroring the Glorot Uniform results in the main text.
Appendix~\ref{app:bhh_liliao_ics} lists the initial conditions of the
discovered orbits, so that the reported orbits can be reproduced
independently.

\section{Forward problem experimental details}
\label{app:forward_details}

This appendix provides full tables, figures, and experimental details for
the forward problem experiments summarized in
Section~\ref{sec:experiments}.

\subsection{Two-body problem: validation}
\label{app:two_body}

As a validation benchmark, we consider the Kepler two-body problem
($m_1 = 0.99$, $m_2 = 0.01$) using a $3 \times 64$ network with hard
constraints, second-order formulation, and the standard two-phase training
protocol (Table~\ref{tab:hyperparameters}). The PINN achieves position
RMSE of $2.12 \times 10^{-7}$ against the numerical
reference, with final training loss $1.79 \times 10^{-8}$.

\subsection{Three-body Euler orbits: first-order vs.\ second-order}
\label{app:euler_orbits}

We consider the simplest Euler collinear configuration with three equal masses
($m_1 = m_2 = m_3 = 1$). Bodies 1 and 2 orbit symmetrically around the stationary
body 3 at the origin. Initial conditions are:
\begin{align}
    &(x_1, y_1, v_{x1}, v_{y1}) = (-0.5, 0, 0, 0.9), \nonumber\\
    &(x_2, y_2, v_{x2}, v_{y2}) = (0.5, 0, 0, -0.9), \label{eq:euler_ic}\\
    &(x_3, y_3, v_{x3}, v_{y3}) = (0, 0, 0, 0). \nonumber
\end{align}

\subsubsection{First-order system (12 equations)}
A $3 \times 128$ network with hard constraints, trained using the standard
two-phase protocol (Table~\ref{tab:hyperparameters}), achieves good agreement
during the first half-period but degrades thereafter. The ``stationary''
body 3 gradually drifts from the origin, indicating convergence failure
(see Table~\ref{tab:euler_comparison}).

\subsubsection{Second-order system (6 equations)}
With identical hyperparameters but the second-order formulation, the
maximum position error drops by approximately two orders of magnitude
(from $1.28 \times 10^{-1}$ to $7.99 \times 10^{-4}$). The trajectories
show excellent agreement with the numerical solution, and the drift of body 3
is reduced.

\begin{table}[H]
\centering
\caption{Comparison of first-order and second-order formulations for Euler orbits.
The second-order formulation achieves approximately two orders of magnitude lower
maximum position error and reduced drift of the ``stationary'' body~3.
All experiments use a $3 \times 128$ network with hard constraints, Adam
($\eta = 10^{-4}$, $2 \times 10^5$ epochs) followed by L-BFGS
($1.5 \times 10^4$ epochs).}
\label{tab:euler_comparison}
\small
\begin{tabular}{lcccc}
\toprule
\textbf{Formulation} & \textbf{Eqs.} & \textbf{Train loss} &
\textbf{Max pos.\ error} & \textbf{Body 3 drift} \\
\midrule
First-order & $12$ & $3.55 \times 10^{-4}$ &
$1.28 \times 10^{-1}$ & $7.06 \times 10^{-4}$ \\
Second-order & $6$ & $2.72 \times 10^{-5}$ &
$7.99 \times 10^{-4}$ & $3.30 \times 10^{-4}$ \\
\bottomrule
\end{tabular}
\end{table}

\noindent The L-BFGS phase reduces the loss substantially further in the
second-order case than in the first-order one, which is consistent with the
second-order system presenting a loss surface more amenable to
quasi-Newton optimization.

\begin{figure}[H]
    \centering
    \begin{subfigure}[b]{0.48\textwidth}
        \centering
        \includegraphics[width=\textwidth]{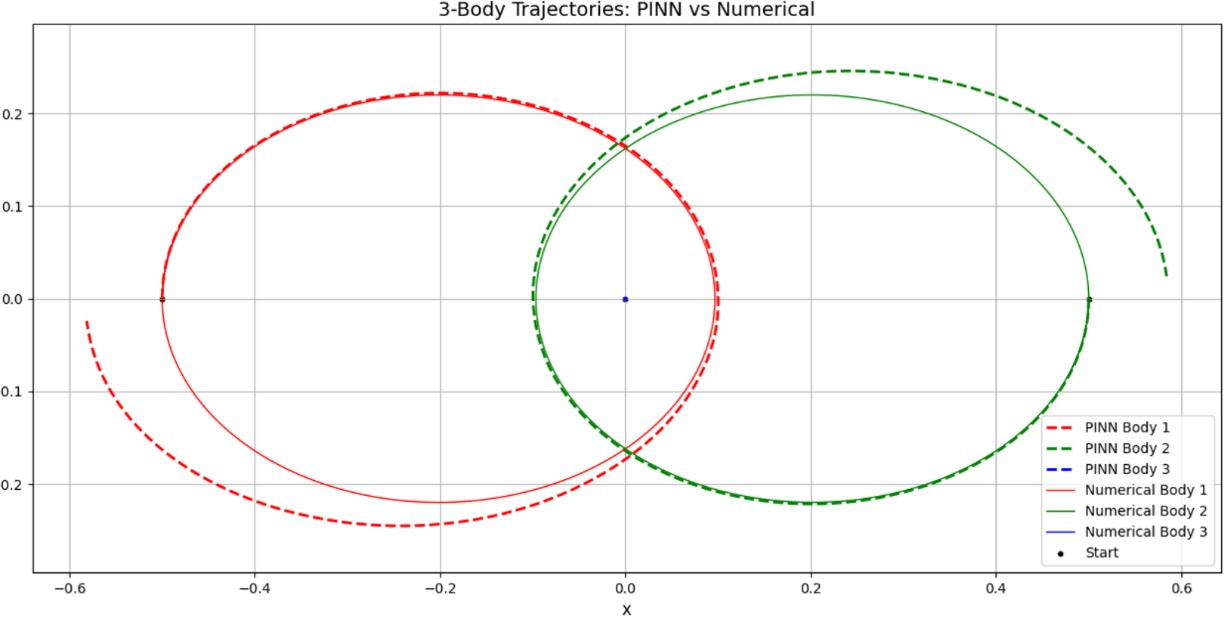}
        \caption{First-order (12 eqs.): body 3 drifts}
    \end{subfigure}
    \hfill
    \begin{subfigure}[b]{0.48\textwidth}
        \centering
        \includegraphics[width=\textwidth]{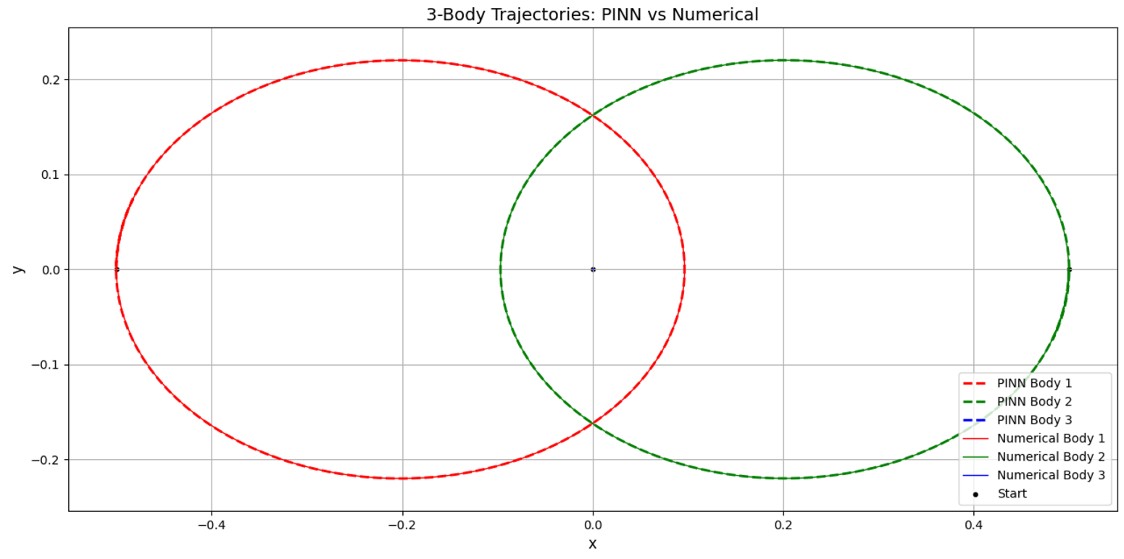}
        \caption{Second-order (6 eqs.): stable}
    \end{subfigure}
    \caption{Euler collinear orbits: PINN trajectories (dashed) vs.\
    numerical reference (solid). The second-order formulation reduces
    maximum position error by ${\sim}\,160\times$
    (Table~\ref{tab:euler_comparison}).}
    \label{fig:euler_comparison}
\end{figure}

\subsection{Three-body Lagrange orbits with RAR}
\label{app:lagrange_orbits}

For Lagrange equilateral triangle orbits, we use three equal masses at the vertices
of an equilateral triangle with side length 2, with angular velocity $\omega = 0.3$
about the center of mass. The period is approximately $T \approx 4.4$ in the original
time scale.

\subsubsection{Importance of time rescaling}
Training in the original time scale ($T \approx 4.4$) fails: inputs exceed the
effective range of $\tanh$ \citep{Theodosiou2026} and the PINN stagnates at
RMSE $\sim 1$. Rescaling to $T \approx 0.5$
yields RMSE $\sim 2 \times 10^{-2}$---an improvement of a factor of ${\sim}\,45$
with identical network parameters. This confirms that time-domain compression
to match the activation function's range is a practical prerequisite for PINNs
on periodic orbits.

\subsubsection{Standard vs.\ modified RAR}
With 64 initial collocation points, L-BFGS achieves low training loss but fails to
generalize: test loss is three to four orders of magnitude higher.
Standard RAR removes the gap but at a worse absolute accuracy
(RMSE $3.26 \times 10^{-2}$). Our modified RAR (60th--95th
percentile sampling) produces steady convergence in both training and test losses,
with trajectories that closely match the numerical solution and even provide reasonable
short-term extrapolation beyond the training interval.

\begin{table}[H]
\centering
\caption{Lagrange orbit accuracy: comparison of sampling strategies.
All experiments use a $3 \times 64$ network, second-order formulation with
hard constraints. ``No RAR'' uses 64 fixed collocation points; standard RAR
adds $n_\text{add} = 64$ points with maximum residual; modified RAR samples from
the 60th--95th percentile range.}
\label{tab:lagrange_rar_comparison}
\small
\begin{tabular}{lccc}
\toprule
\textbf{Strategy} & \textbf{Train loss} & \textbf{Test loss} &
\textbf{RMSE (pos.)} \\
\midrule
No RAR (no scaling) & $2.65 \times 10^{-2}$ & $3.13 \times 10^{-2}$ &
$8.61 \times 10^{-1}$ \\
No RAR (64 pts) & $1.28 \times 10^{-1}$ & $3.10 \times 10^{-1}$ &
$1.93 \times 10^{-2}$ \\
No RAR (64 pts) + L-BFGS & $4.33 \times 10^{-4}$ & $1.31 \times 10^{0}$ &
$1.69 \times 10^{-3}$ \\
Standard RAR & $2.43 \times 10^{-1}$ & $3.01 \times 10^{-1}$ &
$3.26 \times 10^{-2}$ \\
Modified RAR (60--95th pct.) & $4.57 \times 10^{-2}$ & $4.92 \times 10^{-2}$ &
$1.05 \times 10^{-2}$ \\
\bottomrule
\end{tabular}
\end{table}

\begin{figure}[H]
    \centering
    \includegraphics[width=0.7\textwidth]{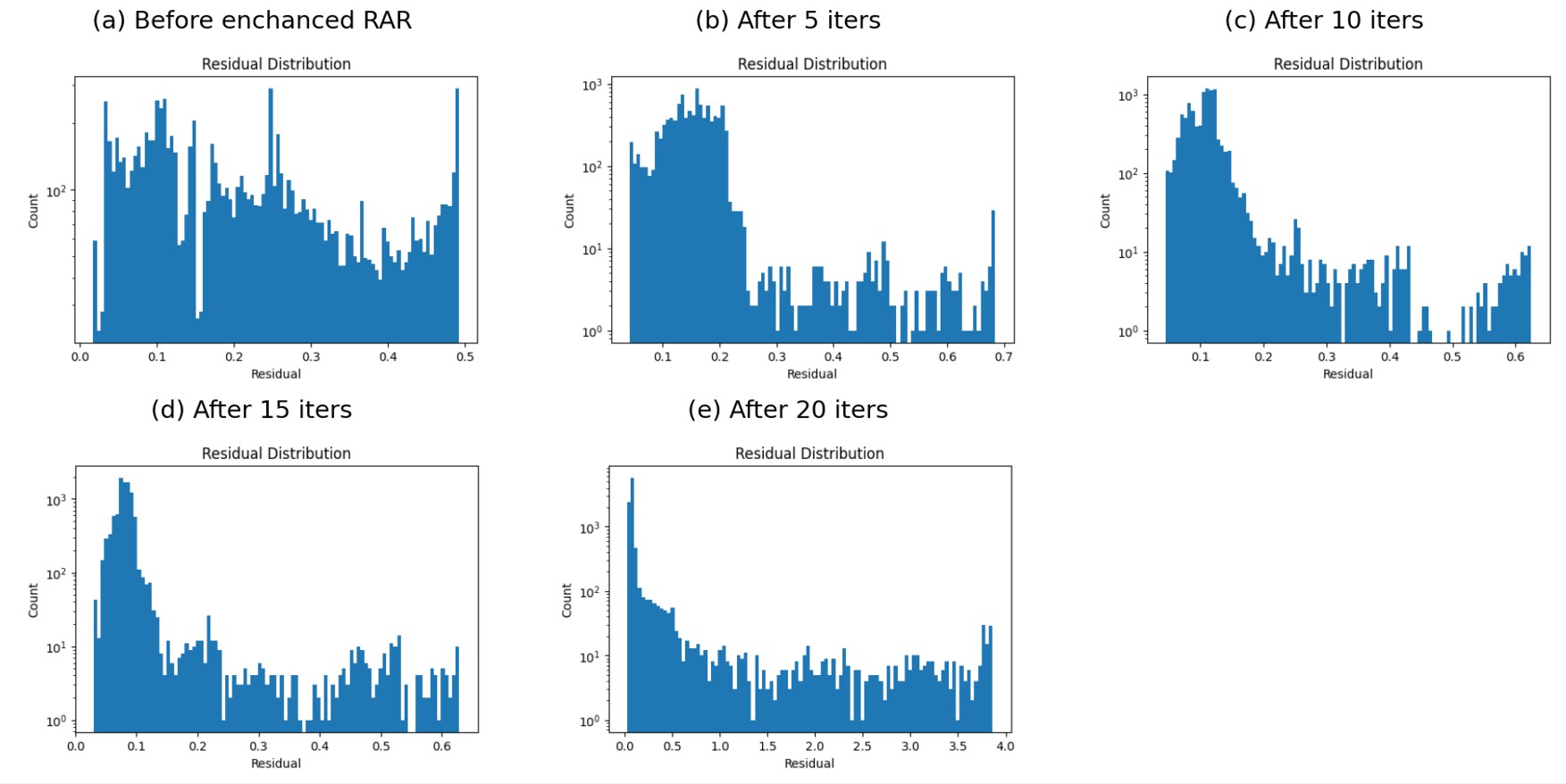}
    \caption{Evolution of residual distributions during modified RAR iterations
    for Lagrange orbits. The 60th--95th percentile sampling strategy produces
    a steady reduction in residual magnitudes across iterations, in contrast to
    standard RAR which concentrates points at extreme outliers.}
    \label{fig:lagrange_rar_hist}
\end{figure}

\begin{figure}[H]
    \centering
    \begin{subfigure}[b]{0.48\textwidth}
        \centering
        \includegraphics[width=\textwidth]{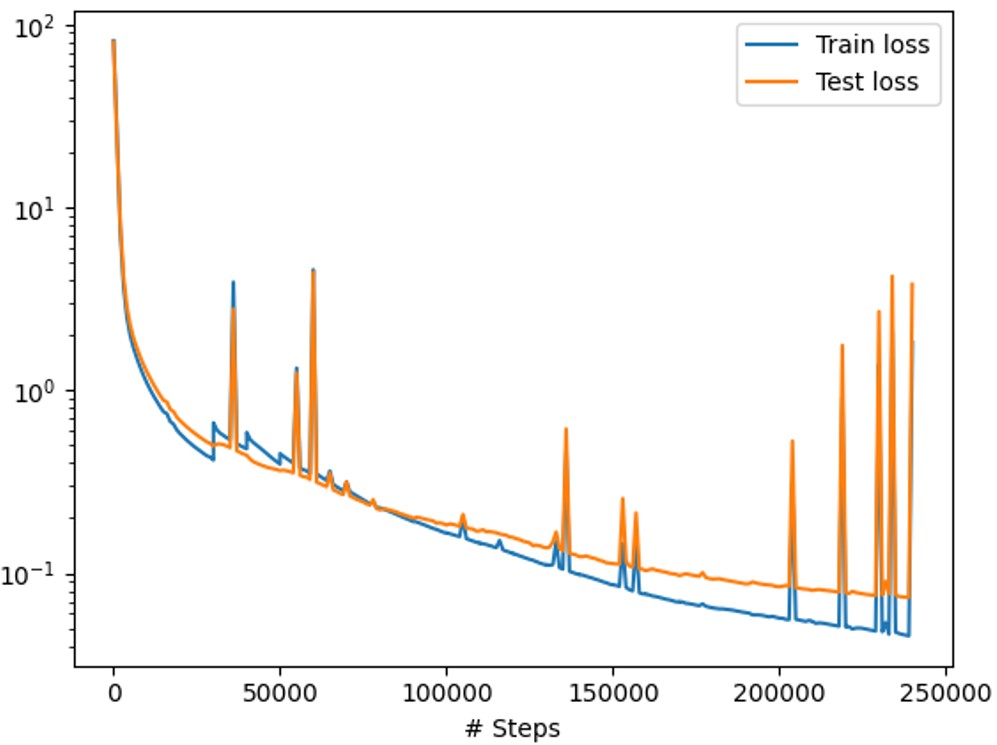}
        \caption{Training and test loss convergence}
    \end{subfigure}
    \hfill
    \begin{subfigure}[b]{0.48\textwidth}
        \centering
        \includegraphics[width=\textwidth]{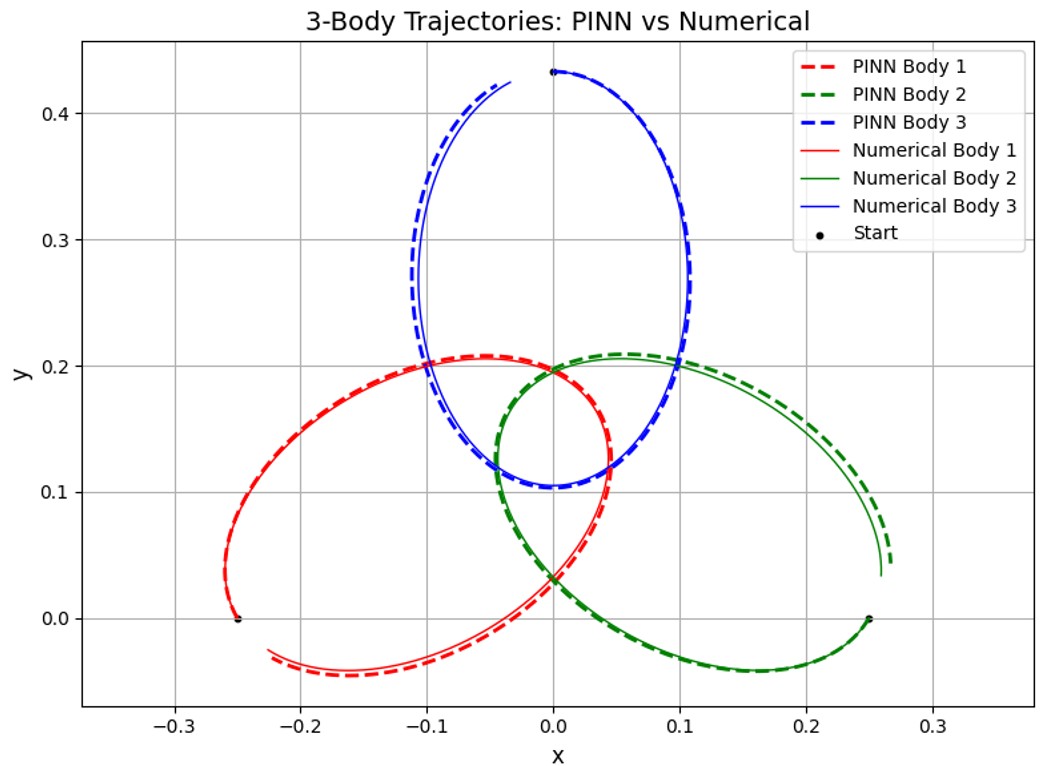}
        \caption{PINN vs.\ numerical trajectories}
    \end{subfigure}
    \caption{Lagrange orbits with modified RAR: (a) Loss convergence showing
    train and test losses decreasing together. (b) Resulting trajectories (dashed)
    closely matching the numerical solution (solid).}
    \label{fig:lagrange_rar}
\end{figure}

\subsection{Figure-eight orbits with Fourier features}
\label{app:figure_eight}

For the figure-eight orbit \citep{Chenciner2000}, initial conditions after time
rescaling to $T = 1$ are used with equal masses. Without the Fourier feature
transformation \eqref{eq:fourier_features}, both $3 \times 64$ and $5 \times 64$
networks fail: training loss decreases while test loss diverges after $\sim 35{,}000$
epochs, and the predicted trajectories bear little resemblance to the true solution.

With the Fourier feature transformation, the $5 \times 64$ network achieves stable
convergence with training and test losses decreasing together. Phase portraits for
both half-period and full-period intervals confirm good agreement with the numerical
solution, with body 1 showing the best accuracy.

The phase-shifted terms give a $2.7\times$ lower RMSE than a Fourier
encoding without them. As shown in Section~\ref{subsec:fourier}, the shifts
add no representational content, so this difference reflects optimization
rather than expressivity.

\begin{table}[H]
\centering
\caption{Figure-eight orbit: effect of Fourier feature transformation on PINN
accuracy. Both experiments use Adam ($\eta = 10^{-4}$, $3 \times 10^5$ epochs)
followed by L-BFGS ($1.5 \times 10^4$ epochs), second-order formulation with
hard constraints.}
\label{tab:figure_eight_comparison}
\small
\begin{tabular}{lcccc}
\toprule
\textbf{Configuration} & \textbf{Net} & \textbf{Train loss} &
\textbf{Test loss} & \textbf{RMSE (pos.)} \\
\midrule
No Fourier & $3{\times}64$ & $7.89 \times 10^{-6}$ &
$1.66 \times 10^{1}$ & $2.58 \times 10^{-1}$ \\
No Fourier & $5{\times}64$ & $1.12 \times 10^{-6}$ &
$7.82 \times 10^{0}$ & $2.50 \times 10^{-1}$ \\
Fourier (no phase) & $5{\times}64$ & $2.13 \times 10^{-5}$ &
$3.78 \times 10^{-5}$ & $4.14 \times 10^{-4}$ \\
Fourier + phase & $5{\times}64$ & $1.27 \times 10^{-5}$ &
$1.59 \times 10^{-5}$ & $1.53 \times 10^{-4}$ \\
\bottomrule
\end{tabular}
\end{table}

\begin{figure}[H]
    \centering
    \begin{subfigure}[b]{0.48\textwidth}
        \centering
        \includegraphics[width=\textwidth]{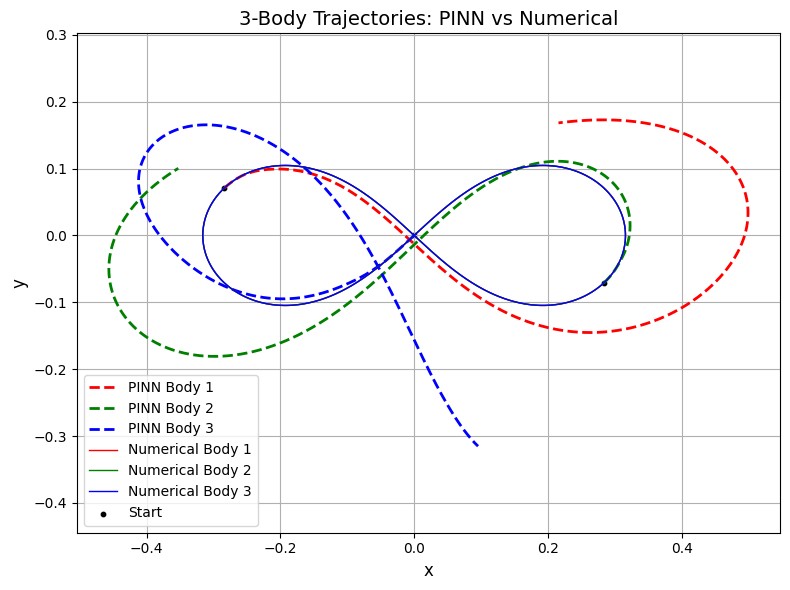}
        \caption{Without Fourier features ($3\times 64$)}
    \end{subfigure}
    \hfill
    \begin{subfigure}[b]{0.48\textwidth}
        \centering
        \includegraphics[width=\textwidth]{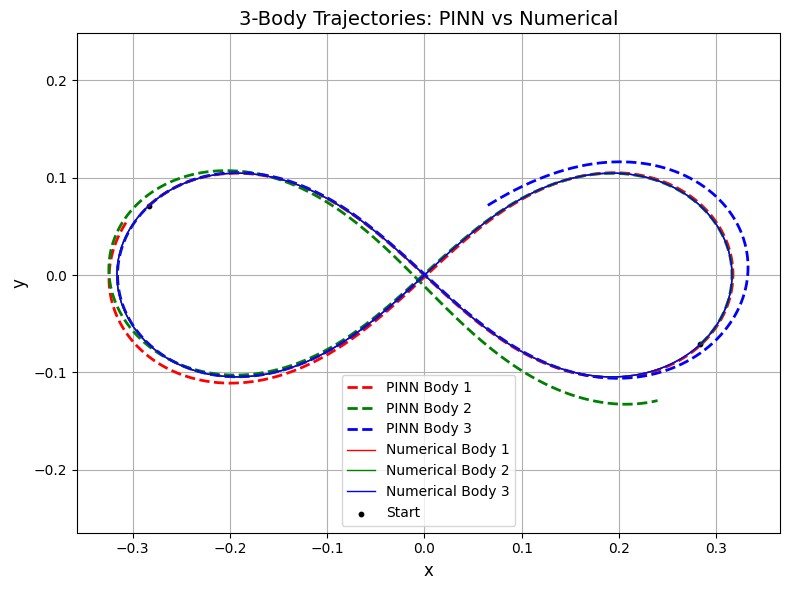}
        \caption{With Fourier features ($5\times 64$)}
    \end{subfigure}
    \caption{Figure-eight orbit trajectories: (a) Standard PINN fails to capture
    the symmetric structure---predicted trajectories diverge from the true solution
    (test loss $\sim 10^{1}$).
    (b) With the Fourier feature input transformation encoding the periodicity and
    $120^\circ$ phase offsets, the PINN closely reproduces the figure-eight orbit
    (RMSE $\sim 10^{-4}$; see Table~\ref{tab:figure_eight_comparison}).}
    \label{fig:figure_eight_traj}
\end{figure}

\subsection{Conservation law diagnostics}
\label{app:conservation}

As an independent validation of solution quality, we compute the total energy
and linear momentum of the PINN solutions and compare their conservation
properties to the numerical reference. For the three-body problem, the total
energy is:
\begin{equation}
    E_{\mathrm{tot}} = \sum_{i=1}^{3} \frac{1}{2} m_i |\dot{\vect{r}}_i|^2
    - G \sum_{i < j} \frac{m_i m_j}{|\vect{r}_i - \vect{r}_j|},
    \label{eq:energy}
\end{equation}
and the total linear momentum is
$\vect{P} = \sum_{i=1}^{3} m_i \dot{\vect{r}}_i$.
Since the PINN does not explicitly enforce conservation laws, any departure
from constant $E_{\mathrm{tot}}$ or $\vect{P}$ provides a physics-based error measure that is
independent of the training loss.

\begin{table}[H]
\centering
\caption{Conservation law diagnostics for forward-problem PINN solutions.
$\Delta E_\text{rel} = \max_t |E(t) - E(0)|/|E(0)|$ measures relative energy
drift; $\Delta P = \max_t |\vect{P}(t) - \vect{P}(0)|$ measures absolute
momentum drift. The numerical reference values ($\texttt{solve\_ivp}$) are
shown for comparison.}
\label{tab:conservation}
\small
\begin{tabular}{lcccc}
\toprule
\textbf{Orbit} & \multicolumn{2}{c}{\textbf{PINN}} &
\multicolumn{2}{c}{\textbf{Numerical ref.}} \\
\cmidrule(lr){2-3} \cmidrule(lr){4-5}
& $\Delta E_\text{rel}$ & $\Delta P$ &
$\Delta E_\text{rel}$ & $\Delta P$ \\
\midrule
Euler & $7.74 \times 10^{-4}$ & $2.65 \times 10^{-4}$ &
$9.99 \times 10^{-10}$ & $0.00$ \\
Lagrange & $3.23 \times 10^{-2}$ & $5.29 \times 10^{-3}$ &
$9.48 \times 10^{-10}$ & $5.41 \times 10^{-15}$ \\
Figure-eight & $1.64 \times 10^{-4}$ & $7.22 \times 10^{-5}$ &
$2.91 \times 10^{-10}$ & $2.67 \times 10^{-15}$ \\
\bottomrule
\end{tabular}
\end{table}

\noindent The PINN solutions conserve energy to within $\sim 0.02$--$3\%$
over one orbital period, which is consistent with the position-level accuracy
reported in Tables~\ref{tab:euler_comparison}--\ref{tab:figure_eight_comparison}.
While this is far from the machine-precision conservation achieved by symplectic
integrators, it provides an independent confirmation that the learned trajectories
are physically meaningful.

% ============================================================
\section{PINN vs.\ standard neural network comparison}
\label{app:pinn_vs_nn}
% ============================================================

\begin{figure}[H]
    \centering
    \begin{subfigure}[b]{0.48\textwidth}
        \centering
        \includegraphics[width=\textwidth]{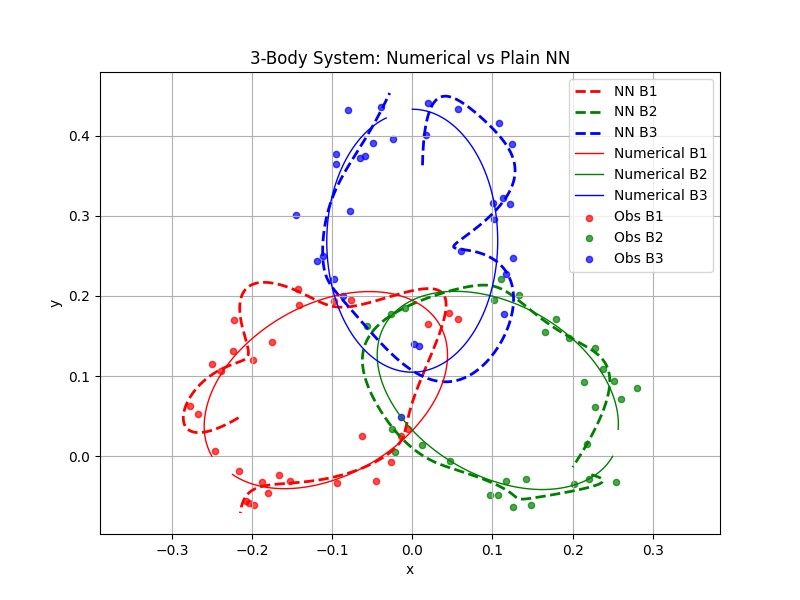}
        \caption{Standard neural network}
    \end{subfigure}
    \hfill
    \begin{subfigure}[b]{0.48\textwidth}
        \centering
        \includegraphics[width=\textwidth]{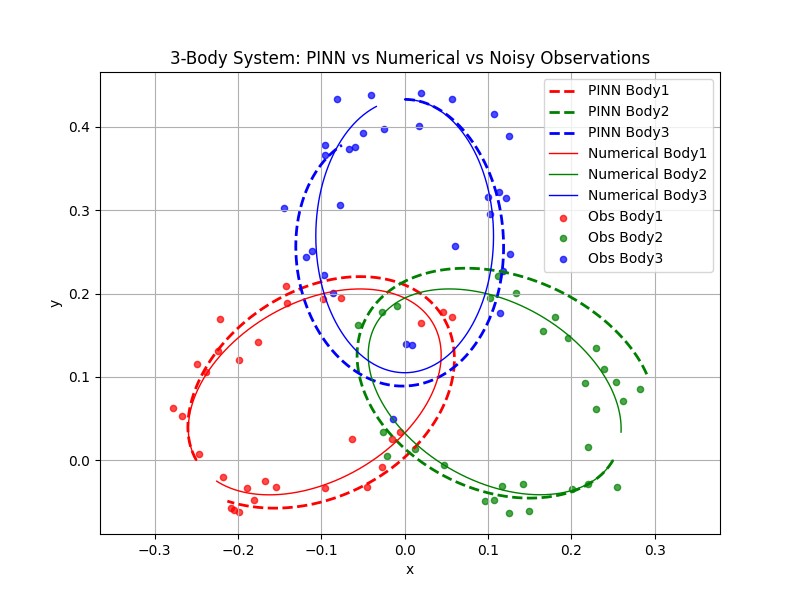}
        \caption{Physics-Informed Neural Network}
    \end{subfigure}
    \caption{Comparison of a standard feedforward network (a) and a PINN with hard
    constraints (b) trained on 90 noisy data points (20\% noise, shown as dots)
    from Lagrange orbits. Both networks achieve comparable RMSE
    (standard network: $2.20 \times 10^{-2}$; PINN: $4.27 \times 10^{-2}$),
    but the standard network produces distorted trajectories, while the PINN
    recovers smooth, physically consistent orbits. The critical difference is
    in conservation: the PINN maintains $\Delta E_\text{rel} = 5.65 \times 10^{-2}$
    and $\Delta P = 7.42 \times 10^{-3}$, compared to $\Delta E_\text{rel} = 5.42$
    and $\Delta P = 4.40$ for the standard network---demonstrating that the ODE
    residual loss acts as a physics-informed regularizer.}
    \label{fig:pinn_vs_nn}
\end{figure}

This appendix establishes the value of the physics constraint for
noisy data. We compare a
standard feedforward network with a PINN (with hard-constrained initial conditions)
trained on 90 noisy data points (20\% noise level) from Lagrange orbits.
The standard network produces distorted trajectories that attempt to interpolate
through all noisy points (Figure~\ref{fig:pinn_vs_nn}a). The PINN, leveraging
the gravitational ODE residual as an inductive bias, recovers smooth, physically
consistent trajectories (Figure~\ref{fig:pinn_vs_nn}b). Additional
training epochs further improve the PINN results while \emph{degrading} the
standard network's performance---consistent with the ODE residual acting as a
regularizer that restricts the hypothesis space to physically plausible
solutions.

% ============================================================
\section{Trainable scaling parameter: detailed results}
\label{app:trainable_c_results}
% ============================================================

This appendix reports the trainable scaling parameter results.
We employ the trainable parameter $C$
(Section~\ref{subsec:trainable_C}) with $C_0 = 3$ and static weights
$w_r = 0.1$, $w_d = 10$. Over $1.5 \times 10^5$ epochs, $C$ increases
approximately linearly (${\sim}\,0.1$ per 1000 epochs after an initial
transient), acting as an implicit annealing schedule
(Eq.~\ref{eq:scaled_residual}): because
$\partial\tilde{\mathcal{L}}_r/\partial C < 0$ always, $C$~grows
monotonically and its final value depends on training duration.

Caveat: the fixed-weight baseline uses $w_d = 40$ versus $w_d = 10$
for the trainable-$C$ configuration; a fully controlled comparison would
hold $w_d$ fixed. The result demonstrates that trainable~$C$ \emph{enables}
orbit discovery, but the improvement magnitude should be interpreted with
this confound in mind.

\begin{table}[H]
\centering
\caption{Effect of the trainable scaling parameter $C$: comparison of
data-driven PINN configurations for Lagrange orbits.
Data: 90 noisy observations (20\% noise), no initial conditions enforced.}
\label{tab:trainable_c_comparison}
\begin{tabular}{lccc}
\toprule
\textbf{Configuration} & \textbf{Final loss} & \textbf{RMSE (position)} &
\textbf{Final $C$ value} \\
\midrule
Fixed $w_r = 0.1$, $w_d = 40$ (no $C$) & $9.47 \times 10^{-1}$ &
$1.85 \times 10^{-1}$ & --- \\
Trainable $C$ ($C_0 = 3$, $1/C^4$) & $4.47 \times 10^{-3}$ &
$7.98 \times 10^{-3}$ & $\approx 17$ \\
Hard constraints (reference) & $3.52 \times 10^{-2}$ &
$1.18 \times 10^{-2}$ & --- \\
\bottomrule
\end{tabular}
\end{table}

\begin{figure}[H]
    \centering
    \begin{subfigure}[b]{0.48\textwidth}
        \centering
        \includegraphics[width=\textwidth]{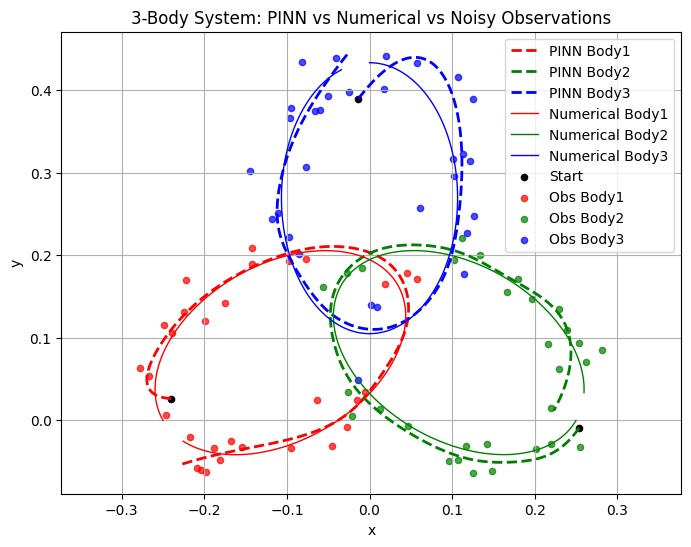}
        \caption{PINN trajectories vs.\ numerical solution}
    \end{subfigure}
    \hfill
    \begin{subfigure}[b]{0.48\textwidth}
        \centering
        \includegraphics[width=\textwidth]{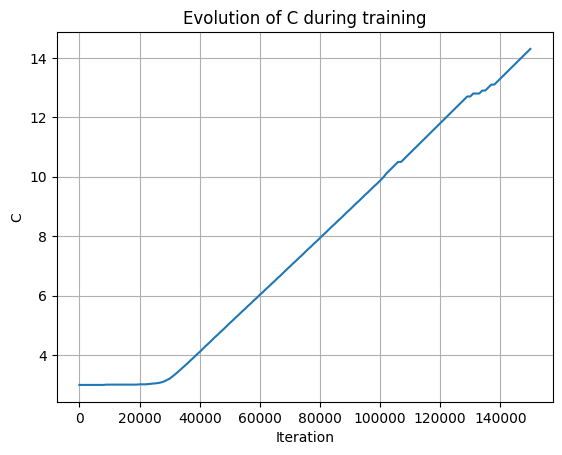}
        \caption{Evolution of $C$ over training}
    \end{subfigure}
    \caption{Data-driven PINN with trainable $C$:
    (a) Trajectories (dashed) vs.\ numerical solution (solid) and noisy data (dots).
    (b) $C$ increases approximately linearly, acting as an implicit annealing schedule.}
    \label{fig:trainable_c}
\end{figure}

% ============================================================
\section{Glorot Normal initialization: detailed results}
\label{app:gn_results}
% ============================================================

This appendix presents the full Glorot Normal (GN) 100-seed experiment,
mirroring the Glorot Uniform results in
Section~\ref{subsec:statistical_robustness}.

\begin{table}[H]
\centering
\caption{Orbit family distribution over $N = 100$ independent seeds
with Glorot Normal initialization and noisy Lagrange training data.
The distribution closely mirrors the Glorot Uniform results
(Table~\ref{tab:seed_results_gu}), with comparable percentages across
all families.}
\label{tab:seed_results_gn}
\footnotesize
\begin{tabular}{lrrcp{5.2cm}r}
\toprule
\textbf{Family} & \textbf{Count} & \textbf{\%} & \textbf{Mean loss}
& \textbf{Distinct $T^*$ values} & \textbf{Refine.\ (\%)} \\
\midrule
Lagrange & 65 & 65 & 0.261 & 6.664 & 5.76 \\
Euler    & 11 & 11 & 0.475 & 7.854 & 2.69 \\
BHH-like & 13 & 13 & 0.381 & 7.327,\; 9.658,\; 19.302 & 26.51 \\
Other    & 10 & 10 & 0.378 & 4.960,\; 7.218,\; 9.686,\; 9.918,\; 9.920,\; 12.097 & 69.45 \\
Figure-eight & 1 & 1 & 0.378 & 9.238 & 54.67 \\
\bottomrule
\end{tabular}
\end{table}

\begin{figure}[H]
    \centering
    \includegraphics[width=0.85\textwidth]{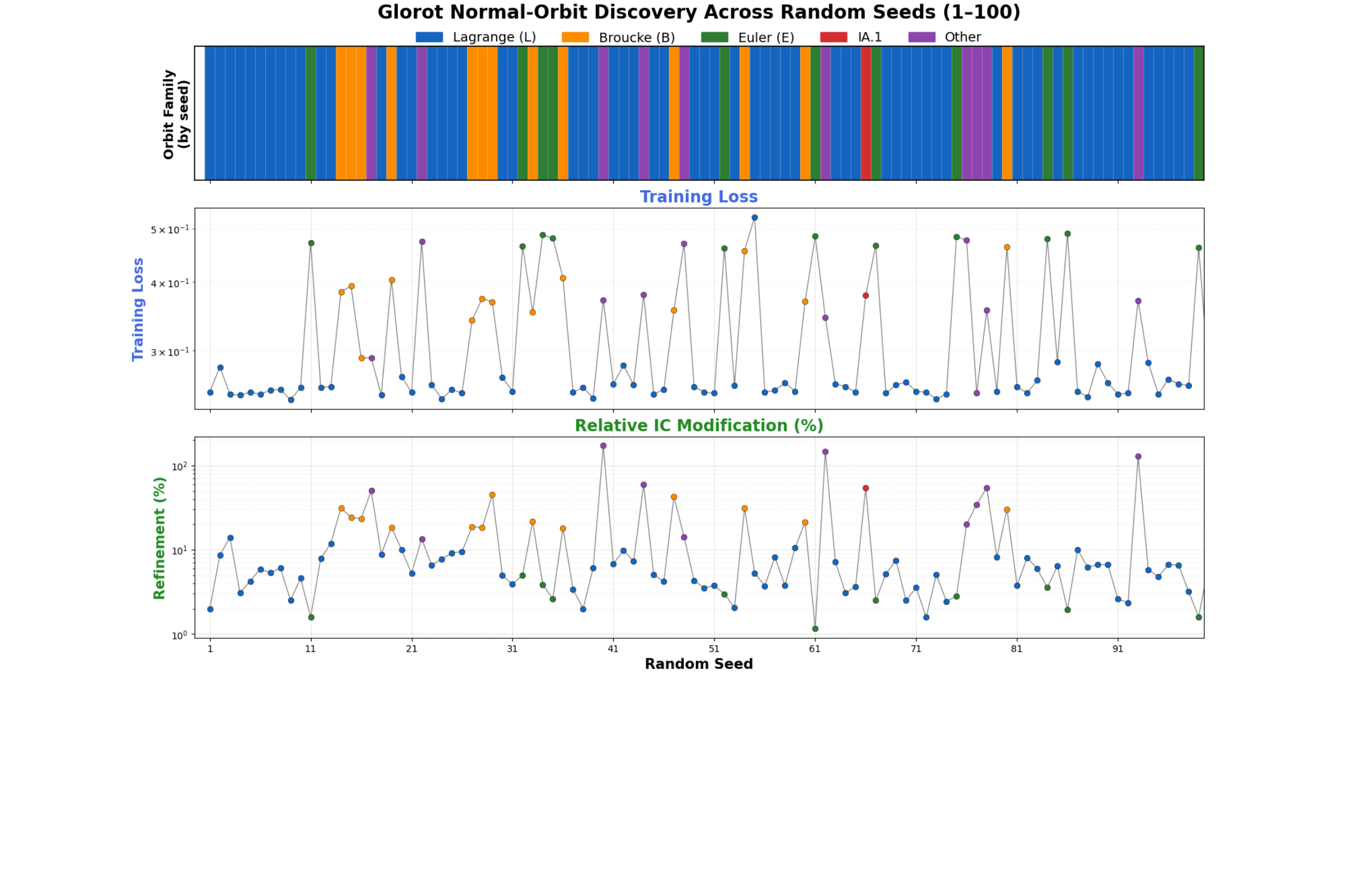}
    \caption{Glorot Normal: orbit family distribution across 100 random seeds,
    with the same layout as Figure~\ref{fig:seed_heatmap}. The overall pattern
    closely resembles the Glorot Uniform experiment, with Lagrange orbits dominating
    and BHH-like, Euler, and ``other'' orbits appearing at comparable frequencies.
    A single figure-eight orbit (red, seed~66) is present.}
    \label{fig:seed_heatmap_gn}
\end{figure}

\begin{figure}[H]
    \centering
    \includegraphics[width=0.65\textwidth]{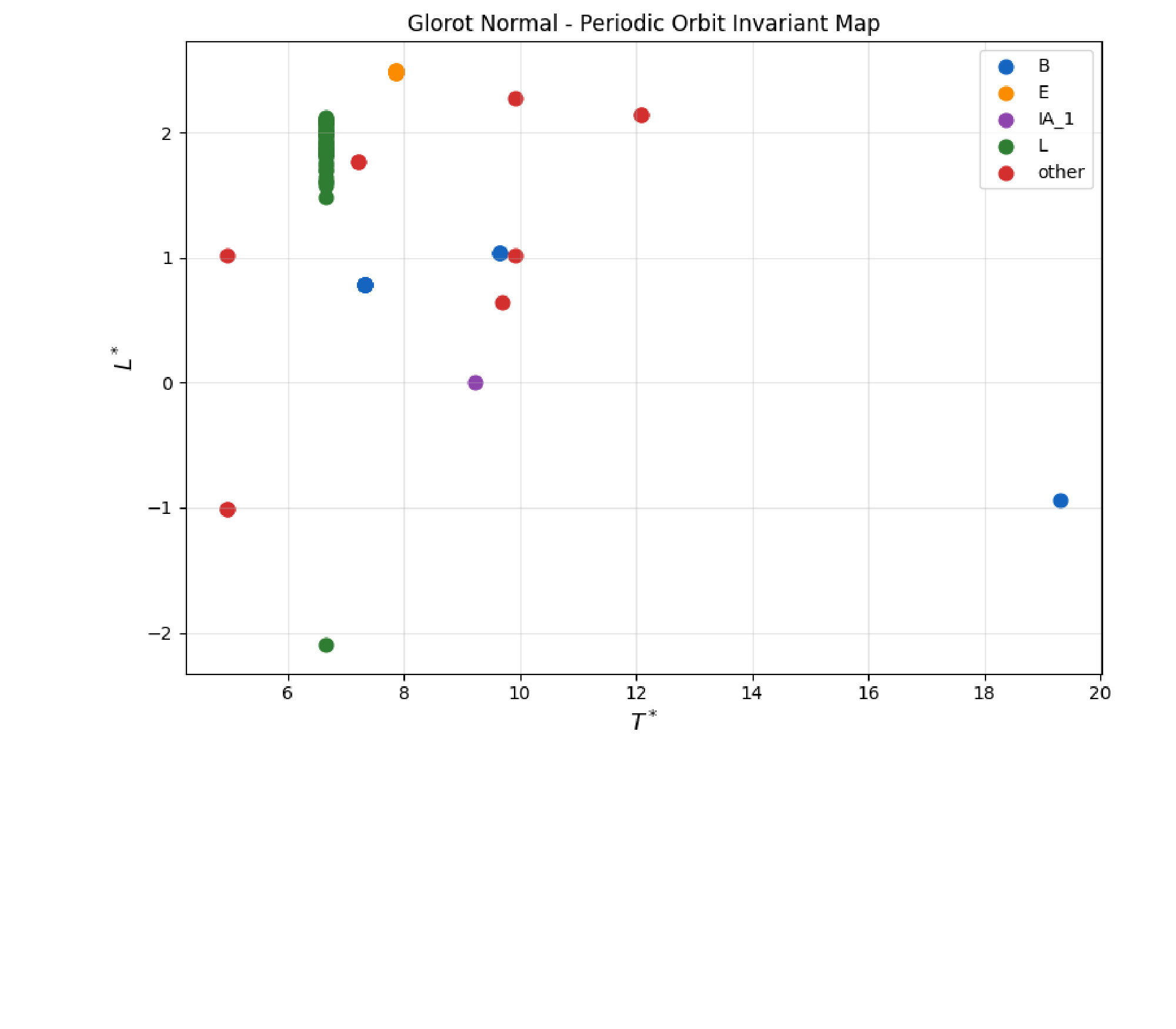}
    \caption{Glorot Normal: periodic orbit invariant map.
    The structure mirrors the GU map (Figure~\ref{fig:orbit_invariant_map}),
    with the addition of an figure-eight orbit (purple) at
    $(T^*, L^*) \approx (9.24, 0)$. The two initialization
    distributions produce distributions of families that the $\chi^2$ test
    cannot distinguish (Table~\ref{tab:chi_squared}).}
    \label{fig:orbit_invariant_map_gn}
\end{figure}

% ============================================================
\section{Discovered orbit initial conditions}
\label{app:bhh_liliao_ics}
% ============================================================

This appendix lists the initial conditions for the two cross-family
orbit discoveries reported in Sections~\ref{subsec:bhh_discovery}
and~\ref{subsec:liliao_discovery}.

\begin{table}[H]
\centering
\caption{PINN-discovered initial conditions for the BHH-type orbit, obtained
from a network trained on 90 noisy figure-eight observations (20\% noise)
without initial conditions. The refined values are obtained via
\texttt{scipy.optimize.least\_squares} starting from the PINN estimates.}
\label{tab:bhh_ics}
\small
\begin{tabular}{lcccc}
\toprule
& $x_i(0)$ & $y_i(0)$ & $\dot{x}_i(0)$ & $\dot{y}_i(0)$ \\
\midrule
\multicolumn{5}{l}{\emph{PINN-inferred (raw)}} \\
Body 1 & $-0.0206$ & $\phantom{-}0.2063$ & $\phantom{-}0.1713$ & $-1.6498$ \\
Body 2 & $\phantom{-}0.2684$ & $\phantom{-}0.0918$ & $\phantom{-}0.9394$ & $\phantom{-}0.6423$ \\
Body 3 & $-0.2236$ & $-0.3013$ & $-1.1534$ & $\phantom{-}1.0108$ \\
\midrule
\multicolumn{5}{l}{\emph{Refined (least-squares, $\delta_T \approx 9.4 \times 10^{-10}$, $T \approx 1.4946$)}} \\
Body 1 & $-0.0070$ & $\phantom{-}0.2174$ & $\phantom{-}0.1896$ & $-1.6361$ \\
Body 2 & $\phantom{-}0.2767$ & $\phantom{-}0.1006$ & $\phantom{-}0.9989$ & $\phantom{-}0.6273$ \\
Body 3 & $-0.2339$ & $-0.2639$ & $-1.1885$ & $\phantom{-}1.0088$ \\
\bottomrule
\end{tabular}
\end{table}

\begin{table}[H]
\centering
\caption{Least-squares refined initial conditions for the figure-eight orbit
discovered by seed~66 (Glorot Normal, Lagrange training data). Period
$T = 1.518234$, invariant period $T^* = 9.237683$, total energy
$E = -3.332852793$, closure error $\delta_T = 1.07 \times 10^{-10}$.}
\label{tab:liliao_ics}
\small
\begin{tabular}{lcccc}
\toprule
& $x_i(0)$ & $y_i(0)$ & $\dot{x}_i(0)$ & $\dot{y}_i(0)$ \\
\midrule
Body 1 & $-0.1428$ & $\phantom{-}0.3893$ & $-0.7716$ & $-0.1352$ \\
Body 2 & $\phantom{-}0.1514$ & $-0.0753$ & $\phantom{-}1.5879$ & $-1.1400$ \\
Body 3 & $-0.0080$ & $-0.3142$ & $-0.8163$ & $\phantom{-}1.2752$ \\
\bottomrule
\end{tabular}
\end{table}

\end{document}